%% file: main.tex
\documentclass[10pt,twocolumn,letterpaper]{article}

\usepackage{cvpr}              

\input{preamble}

\definecolor{cvprblue}{rgb}{0.21,0.49,0.74}
\usepackage[pagebackref,breaklinks,colorlinks,allcolors=cvprblue]{hyperref}

\def\paperID{3260} 
\def\confName{CVPR}
\def\confYear{2026}

\title{VIVA: VLM-Guided Instruction-Based Video Editing with Reward Optimization}

\author{
    Xiaoyan Cong$^{1, 2, *}$, Haotian Yang$^{2, \dagger}$, Angtian Wang$^{2}$, Yizhi Wang$^{2}$, Yiding Yang$^{2}$, Canyu Zhang$^{2}$\\
    Chongyang Ma$^{2}$ \\
    $^{1}$ Brown University \quad $^{2}$ Intelligent Creation, ByteDance
}

\begin{document}

\input{figures/fig_teaser}

\maketitle

\renewcommand{\thefootnote}{\fnsymbol{footnote}} 

\footnotetext[1]{Work done while Xiaoyan Cong was an intern at ByteDance.} 
\footnotetext[2]{Project Lead.}

\renewcommand{\thefootnote}{\arabic{footnote}}

\input{sec/0_abstract}    
\input{sec/1_introduction}
\input{sec/2_related_work}
\input{sec/5_method}
\input{sec/6_experiment}

\input{sec/8_conclusion}
{
    \small
    \bibliographystyle{ieeenat_fullname}
    \bibliography{main}
}

\clearpage
\maketitlesupplementary
\appendix
\input{appendix/preliminary}
\input{appendix/implementation_details}
\input{appendix/data_strategy}
\input{appendix/experiments}

\input{appendix/limitation}

\end{document}

%% file: preamble.tex
\usepackage{xspace}
\usepackage{booktabs}    
\usepackage{graphicx}   
\usepackage{multirow}   
\usepackage{colortbl}  
\usepackage{xcolor}     
\usepackage{pifont}    
\usepackage{makecell}    

\definecolor{oursrow}{RGB}{242,247,242}
\definecolor{mygray}{RGB}{100,100,100} 
\newcommand{\prompt}{\textbf{t}}
\newcommand{\token}{\textbf{x}}
\newcommand{\vlmtoken}{\token_{vlm}}
\newcommand{\srctoken}{\token_{src}}
\newcommand{\noisetoken}{\token_{noise}}
\newcommand{\videotoken}{\token_{video}}
\newcommand{\tokenrefiner}{\mathcal{R}}
\newcommand{\projector}{\mathcal{P}}
\newcommand{\patchify}{\mathbf{P}}
\newcommand{\latent}{\mathbf{z}}
\newcommand{\noiselatent}{\latent_{noise}}
\newcommand{\srcvideolatent}{\latent_{src}}
\newcommand{\sourceprompt}{\prompt_{src}}
\newcommand{\editedprompt}{\prompt_{edit}}
\newcommand{\instruction}{\prompt_{ins}}
\newcommand{\image}{\textbf{I}}
\newcommand{\sourceimage}{\image_{src}}
\newcommand{\video}{\textbf{V}}
\newcommand{\sourcevideo}{\video_{src}}

\newcommand{\refimage}{\image_{ref}}
\newcommand{\clipsimilarity}{C}
\newcommand{\rewardscore}{\textbf{R}}
\newcommand{\weight}{w}
\newcommand{\rewardweight}{\weight}
\newcommand{\maskweight}{\weight}
\newcommand{\loss}{\mathcal{L}}

\newcommand{\modelname}{VIVA\xspace}


%% file: figures/fig_teaser.tex
\twocolumn[{%
\renewcommand\twocolumn[1][]{#1}%
\maketitle
\begin{center}
    \captionsetup{type=figure}
   \includegraphics[width=1\linewidth]{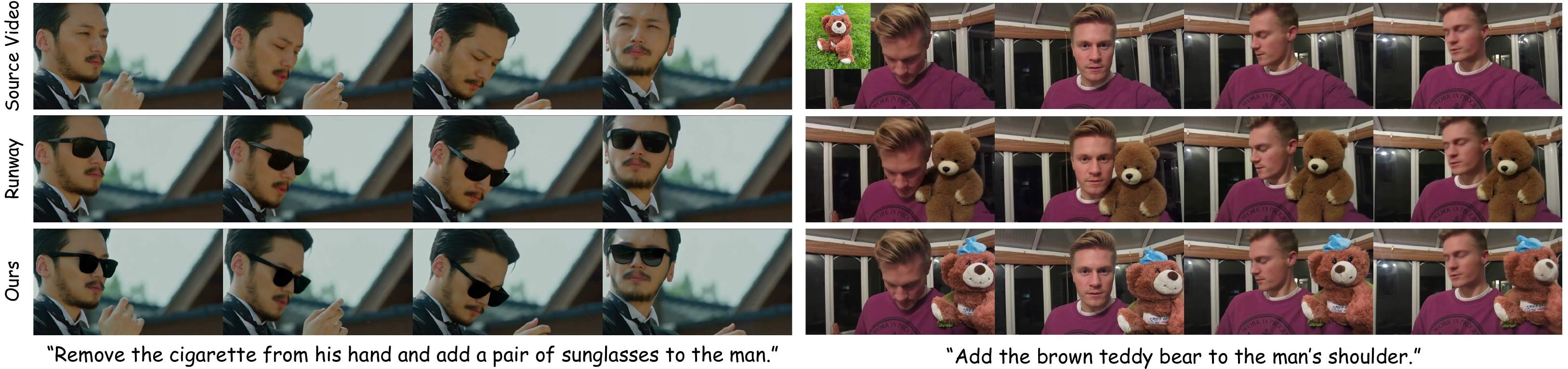}
    \captionof{figure}{
    Example results generated by \modelname in comparison with Runway Gen-4 Aleph~\cite{runway2025aleph}. Our method supports instruction-based video editing with an optional reference image as input (shown as the inset teddy bear in the first row).
    Runway Gen-4 Aleph over-completes the editing instruction, removing both the hand and the cigarette entirely, and fails to preserve the identity of the teddy bear.
    }
    \label{fig:teaser}
    \end{center}%
}]

%% file: sec/0_abstract.tex
\begin{abstract}
Instruction-based video editing aims to modify an input video according to a natural-language instruction while preserving content fidelity and temporal coherence. However, existing diffusion-based approaches are often trained on paired data of simple editing operations, which fundamentally limits their ability to generalize to diverse and complex, real-world instructions. To address this generalization gap, we propose VIVA, a scalable framework for instruction-based video editing that leverages VLM-guided encoding and reward optimization. First, we introduce a VLM-based instructor that encodes the textual instruction, the first frame of the source video, and an optional reference image into visually-grounded instruction representations, providing fine-grained spatial and semantic context for the diffusion transformer backbone. Second, we propose a post-training stage, Edit-GRPO, which adapts Group Relative Policy Optimization to the domain of video editing, directly optimizing the model for instruction-faithful, content-preserving, and aesthetically pleasing edits using relative rewards. Furthermore, we propose a data construction pipeline designed to synthetically generate diverse, high-fidelity paired video–instruction data of basic editing operations. Extensive experiments show that VIVA achieves superior instruction following, generalization, and editing quality over state-of-the-art methods.
Website: \href{https://viva-paper.github.io/}{https://viva-paper.github.io/}.
\end{abstract}

%% file: sec/1_introduction.tex
\section{Introduction}
\label{sec:introduction}

Video editing plays a central role in modern digital media creation, enabling applications such as film post-production, advertising, and language-guided content personalization~\cite{Jiang2025VACE, zhao2023controlvideo, couairon2023videdit, liao2025iclunpaired, runway2025aleph}. Recent diffusion-based approaches typically formulate video editing as a supervised translation problem from an input video and an editing instruction to an edited output. This formulation relies on carefully constructed paired datasets of source videos and corresponding edited targets or text prompts~\cite{zi2025se, zhao2023controlvideo,couairon2023videdit,chen2024eve, ju2025editverse, decart2025lucyedit}.

However, the practical construction of such paired training data faces a significant bottleneck: existing synthesis pipelines are largely confined to generating pairs for highly simplified editing tasks. These are often limited to basic operations like single object addition, deletion, or object swaps strictly within a predefined mask~\cite{zi2025se, cheng2024insv2v, zhao2023controlvideo, chen2024eve}. It remains extremely difficult to automatically generate training pairs for more general and complex edits, such as combining multiple tasks, performing object swaps outside of the original mask, changing the global environment (e.g., weather or season), or executing other semantically-aware controls.

This reliance on training data comprised of simplistic edits fundamentally limits the generalization capabilities of current video editing systems. When faced with open-domain or complex real-world scenarios, their performance degrades. To address this generalization gap, we introduce a method to improve model performance on complex editing tasks, even when trained only on limited, simple editing pairs. Specifically, we introduce two complementary components that enhance how instructions are represented and how the video editing model is optimized for output quality.


First, we introduce the \emph{VLM instructor}, a vision-language model used as a multimodal instruction encoder. Given the first frame of the input video together with the user's editing instructions, the VLM produces a sequence of grounded, fine-grained multimodal tokens that jointly encode the target entities, spatial regions, and desired attribute changes. These tokens are then fed as conditioning signals into the diffusion transformer (DiT) backbone, in place of standard text embeddings. Recent large vision-language models demonstrate strong capabilities in aligning visual content with detailed natural-language instructions and conversations~\cite{liu2023visualinstruction,maaz2024videochatgpt}, and we leverage this capacity to obtain richer, disambiguated instruction representations. This approach allows us to effectively combine the advanced visual-linguistic understanding of the VLM with the high-fidelity video generation capability of the pretrained DiT. 


Second, we introduce \emph{Edit-GRPO}, a post-training stage that explicitly optimizes the model toward successful edits via reinforcement learning (RL). Building on Group Relative Policy Optimization (GRPO)~\cite{shao2024deepseekmath, liu2025flowgrpo}, which has been shown to improve reasoning ability in large language models by comparing groups of sampled outputs under task-specific rewards, we adapt the same principle to text-guided video editing. For each input video and carefully designed editing instruction, the model generates multiple candidate edited videos; we then compute edit-specific rewards that measure instruction fidelity, identity preservation, and structural consistency, and use the relative scores within each group to update the policy. This edit-centric RL phase focuses learning on semantic correctness rather than pixel-level reconstruction alone. Previous methods for image editing have employed iterative synthesis, where a model is trained, used to generate new pairs, and then good pairs are labeled and added back to the training set~\cite{shi2024seededit}. This process can be formulated as rejection-with-resampling (RWR)~\cite{peng2019advantage, lee2023aligning} in an RL paradigm. In contrast, our approach directly uses an online RL method GRPO, which more effectively extracts supervision from each limited paired example to achieve better editing performance.


Furthermore, we propose a pipeline to construct a high-quality, synthetic dataset. This dataset consists of paired videos (original and edited) accompanied by accurate text instructions. Our pipeline is designed to generate video pairs demonstrating strict local editing and high fidelity, which cover a diverse range of basic editing operations, providing a solid foundation for training our model.


In summary, our contributions include:
\begin{itemize}
\item The VLM instructor that jointly processes the first frame and text to provide visually-grounded instruction representations for the video editor.
\item A post-traning stage, termed Edit-GRPO, that directly rewards successful, instruction-faithful edits.
\item A data construction pipeline that generates diverse, high-quality paired video–instruction dataset of basic editing operations.
\item Extensive evaluations showing superior instruction following without degrading video quality compared to strong baselines.
\end{itemize}

%% file: sec/2_related_work.tex
\section{Related Work}
\label{sec:related_work}

\paragraph{Instruction-based video editing.}
Natural language–driven video transformation enables intuitive and controllable edits while maintaining visual fidelity.
Earlier methods~\cite{qi2023fatezero, cong2023flatten, wu2023tune, liu2024video, geyer2023tokenflow, kara2024rave, chen2024eve, li2024vidtome, fan2024videoshop, wang2025videodirector, ku2024anyv2v} leveraged inversion-based approaches~\cite{song2020denoising} to perform zero-shot or one-shot text-driven edits on videos.
With the advent of open-source pretrained video diffusion models~\cite{hong2022cogvideo, kong2025hunyuanvideosystematicframeworklarge, wan2025wan}, a wave of task-specific solutions emerged for video addition~\cite{zhuang2025get, tu2025videoanydoor}, inpainting~\cite{zi2025cococo, bian2025videopainter}, virtual try-on~\cite{fang2024vivid, zuo2025dreamvvt, Karras_FashionVDM_2024}, and character replacement/animation~\cite{cheng2025wan}. 
Although effective in limited scenarios, these approaches typically depend on manual priors such as masks, poses, or motion trajectories, which limit flexibility and generalization.
InsV2V~\cite{cheng2024insv2v} pioneers instruction-only video editing by utilizing InstructPix2Pix~\cite{brooks2023instructpix2pix} to synthesize paired video editing data.
However, its results were hindered by the limited data quality and backbone performance.
More recent efforts~\cite{Jiang2025VACE, ju2025editverse, liao2025iclunpaired, decart2025lucyedit, runway2025aleph, ye2025unic, Yu2025VEGGIE} have scaled instruction-based editing with notable gains.
Ditto~\cite{Bai2025Ditto} supplies a pretrained backbone with a series of trainable context-blocks.
Concurrent works like InstructX~\cite{mou2025instructx}, Omni-Video~\cite{tan2025omni}, OmniV2V~\cite{liang2025omniv2v}, and Univideo~\cite{wei2025univideo} also harness VLM's understanding capacities to aid video editing, demonstrating impressive progress.

\paragraph{VLMs for visual understanding and generation.}
Vision–language models (VLMs) provide a bridge between visual perception and textual description.
Foundational VLMs such as CLIP~\cite{radford2021clip}, BLIP-2~\cite{li2023blip2}, and Flamingo~\cite{alayrac2022flamingo} learn joint representations that align visual content with textual descriptions. Instruction-tuned variants, including LLaVA~\cite{liu2023llava}, Video-ChatGPT~\cite{maaz2024videochatgpt}, and Qwen2.5-VL~\cite{bai2025qwen2}, further enhance visual grounding and context-aware textual comprehension. These advances show that pretrained VLMs can serve as powerful priors for conditional control in video generation. Building on this foundation, our work adopts a VLM-based instructor to encode both textual instructions and visual context, improving controllability and instruction fidelity in video editing.

\input{figures/fig_pipeline}

\paragraph{Reinforcement learning for generative models.}
Reinforcement learning has emerged as an effective approach for enhancing model alignment and controllability. 
Algorithms such as PPO~\cite{schulman2017ppo}, RPO~\cite{rahman2022rpo}, and GRPO~\cite{shao2024deepseekmath} have been widely adopted in large language models to refine instruction-following and preference alignment. 
Recent studies~\cite{liu2025videoalign,liu2025flowgrpo,xue2025dancegrpo,wang2025unifiedreward} have extended these ideas to visual generation, while other works~\cite{wu2025rewarddance,luo2025editscore} focus on image editing, employing reward-guided fine-tuning to better align generative edits with user intent. 
We extend these approaches for video editing to comprehensively enhance overall performance, robustness, and generalization. 

%% file: figures/fig_pipeline.tex
\begin{figure*}
\centering
\includegraphics[width=0.98\textwidth]{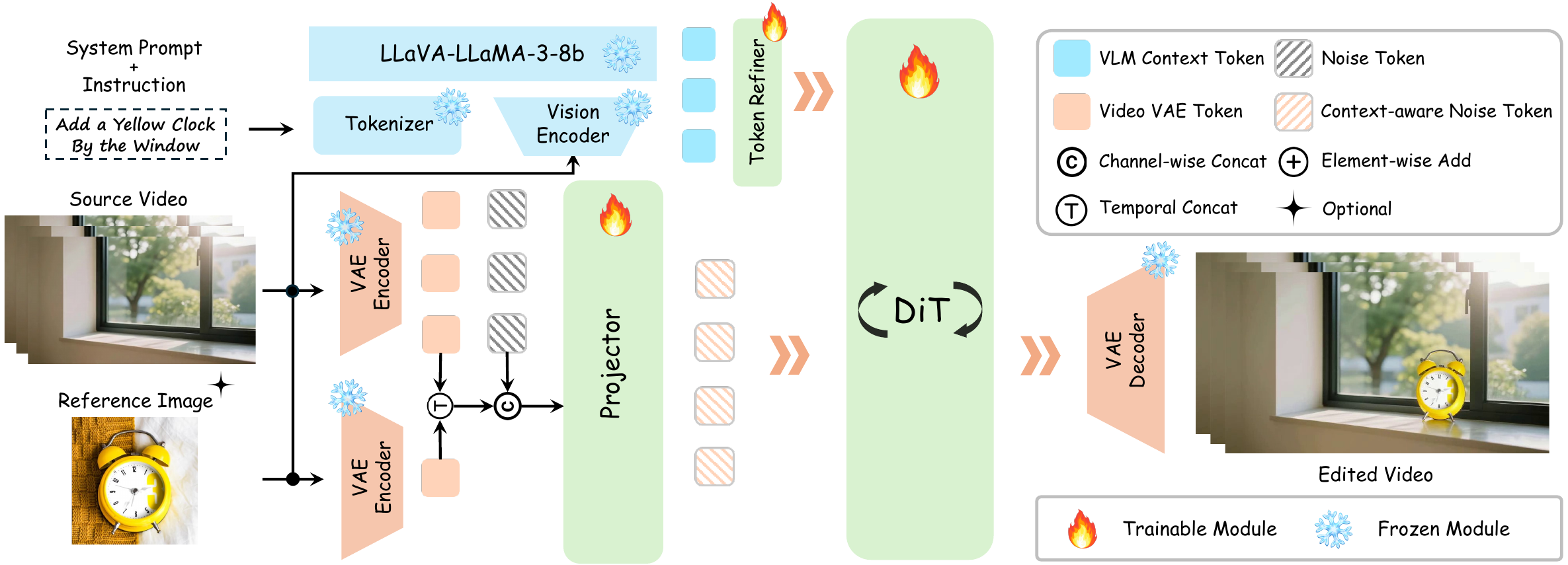}
\caption{Overall pipeline of \modelname. 
A context-aware VLM instructor encodes the system prompt, instruction, first frame of the source video, and an optional reference image into VLM tokens.
A trainable token refiner aligns these tokens to the pretrained DiT latent space. 
The VAE encodings of the source video and optional reference image are added to the noisy latent to form context-aware noise tokens.
Finally, the DiT denoises these tokens under VLM guidance to generate the edited video.
}
\label{fig:pipeline}
\end{figure*}

%% file: sec/5_method.tex
\section{Method}
\label{sec:method}

As illustrated in Figure~\ref{fig:pipeline}, \modelname employs a Diffusion Transformer (DiT)~\cite{kong2025hunyuanvideosystematicframeworklarge} as the generation branch and a Vision-Language Model (VLM) instructor as the understanding branch. 
Section~\ref{subsec:incontext_multimodality_understanding} describes how the VLM instructor aggregates and interprets diverse multimodal conditions, including a text instruction, a source video, and an optional reference image.
Section~\ref{subsec:instruction_based_sft} introduces how \modelname adapts and fine-tunes a pretrained text-to-video (T2V) DiT to perform versatile instruction-based video editing using the semantic context provided by the VLM instructor and the source video.
To enhance instruction following, content fidelity, and visual aesthetics, we introduce a post-training strategy based on Group Relative Policy Optimization (GRPO), termed \textit{Edit-GRPO}, in Section~\ref{subsec:posttraining_grpo}.
Section~\ref{subsec:data_pipeline} presents our data preparation process that generates diverse and high-quality video editing dataset.

\subsection{Context-Aware VLM Instructor}
\label{subsec:incontext_multimodality_understanding}
Since video editing is a highly context-aware task, it is critical and non-trivial to interpret the complicated relationship between diverse multimodal conditions among text instructions, source videos, and optionally provided reference images.
We argue that an encoder exposed only to the language space, such as T5 series~\cite{raffel2020exploring}, struggles to provide sufficient semantic guidance, thereby introducing more difficulty and ambiguity to the subsequent generation branch.
Instead, VLM achieves more robust alignment between language space and visual feature space, making it inherently a great alternative for understanding and processing multimodal context.
There have been several works trying to connect a VLM such as QWen2.5-VL~\cite{bai2025qwen2} with a DiT pretrained on T5 space such as Wan~\cite{wan2025wan}.
However, such strategy requires a heavy pretraining stage for aligning the feature space of the VLM with the generation space of the DiT.
Thus, we choose Hunyuan-T2V-13B~\cite{kong2025hunyuanvideosystematicframeworklarge} as our backbone, as its pretrained model has already aligned with the VLM space (llava-llama-3-8b~\cite{2023xtuner}).

Solely relying on an input of semantically sparse textual instructions $\instruction$ often leads to a less underspecified conditioning signal, causing ambiguity for the subsequent editing process. 
To address this limitation, we propose augmenting the input by incorporating the first frame $\sourceimage$ sampled from the source video $\sourcevideo$. 
The introduction of $\sourceimage$ significantly enriches the condition tokens with grounded, fine-grained semantic bias. 
Consequently, the VLM is better equipped to interpret the intricate relationship between the editing instruction and the source video. 
This integrated understanding allows the VLM to implicitly distinguish critical elements, such as the intended editing region and the specific editing entity. 
By providing the DiT with such fine-grained and accurate control signals, our approach effectively overcomes the ambiguity caused by the sparse conditioning inherent in instruction-only designs, ultimately yielding higher fidelity and more controllable video editing.
To further support a broader range of multimodal control and enhance the user's flexibility to specify the exact editing outcome, our VLM instructor allows an optional reference image input $\refimage$. 
While the editing instruction dictates what transformation should occur, the reference image explicitly grounds how the edited content should look.
This additional visual context takes more advantages of VLM's multimodal understanding and thinking abilities to generate more detailed and accurate editing guidance.

We take the last hidden states from the VLM instructor as our multimodal condition tokens $\vlmtoken$:
\begin{equation}
    \vlmtoken = \text{VLM}(\instruction, \sourceimage, \refimage).
\end{equation}

\subsection{Instruction-based Supervised Fine-Tuning}
\label{subsec:instruction_based_sft}
To effectively preserve the multimodal understanding and reasoning capabilities inherent in the VLM, we freeze its parameters throughout the training process, without losing its strong generalization ability. 
While the pretrained DiT weights of the HunyuanT2V model are already strongly aligned with the VLM's language space, the introduction of additional visual information $\sourceimage$ and $\refimage$ requires a careful adaptation strategy. 
In addition to finetuning the DiT, we introduce and train a lightweight Token Refiner $\tokenrefiner$ to enhance the DiT's comprehensive responsiveness to the visual conditioning signals, as well as the understanding of the underlying connections between textual and visual conditioning information.

To enhance the editing consistency with $\sourcevideo$, we concatenate it with the noise in the channel dimension.
Such design intuitively and effectively introduces a spatially and temporally aligned inductive bias, thereby providing the denoising process with rich and strong contextual guidance of the structure and motion of the source video.
Specifically, a VAE latent $\srcvideolatent$ of $\sourcevideo$ and a noise latent $\noiselatent$ are patchified and unfolded into a 1D sequence of tokens $\srctoken \in \mathbb{R} ^{L \times C}$ and $\noisetoken \in \mathbb{R} ^{L \times C}$ seperately.
These two sequence of tokens are then channel-concatenated to get $[\srctoken,\noisetoken] \in \mathbb{R} ^{L \times 2C} $.
Then the context-aware noisy video tokens can be formulated as
\begin{equation}
    \videotoken = \projector(\text{Concat}(\patchify(\srcvideolatent), \patchify(\noiselatent))^c),
\end{equation}
where $\videotoken \in \mathbb{R} ^{L \times C} $, $\text{Concat}(\cdot, \cdot)^c$ represents channel-wise concatenation, $\patchify$ is a patchification module, and $\projector$ indicates a trainable projector that aggregates the contextual video information along the channel dimension to align with the feature dimension of the DiT.

With the multimodal condition tokens $\vlmtoken$ and context-aware video tokens $\videotoken$, we use Flow Matching~\cite{lipman2022flow} for model training.
In this supervised fine-tuning stage, we train the patchify module $\patchify$, the projector $\projector$, the token refiner $\tokenrefiner$, and the entire DiT on the synthetic paired source/target-video-instruction dataset constructed by ourselves.
Due to the inherent scarcity of paired video editing data and the limited range of editing types, we incorporate large-scale image editing data as an supplement through treating an image as a video with one frame. 
Experiments demonstrate that various editing capabilities within the image domain have effectively been transferred to the video domain, such as global stylization editing, which does not exist in our paired video data.

With the mask video $M$ generated during the data preparation introduced in Section~\ref{subsec:data_pipeline}, we augment the vanilla Rectified Flow loss into a masked version as:
\begin{equation}
    \loss_{\mathrm{mask}} = (\mathbf{1} + \maskweight_{m} M ) \loss_{\mathrm{FM}},
\label{eq:maskloss}
\end{equation}
where $\maskweight_{m}$ is the mask weight.

\subsection{Edit-GRPO}
\label{subsec:posttraining_grpo}
To further enhance model performance, we design a sophisticated reward system via GRPO, focusing on improving the instruction following ability, source video preservation, and human preference alignment.
The overall pipeline of Edit-GRPO is presented in Figure~\ref{fig:pipeline_grpo}.
\input{figures/fig_grpo}

\paragraph{Instruction following.}
The accurate and high-quality instruction following ability is the most fundamental metric for evaluating the performance of an instruction-based video editing model. 
However, since a direct reward model to quantify this does not yet exist, we design the objective as $(\clipsimilarity(\video_{edit}, \editedprompt) - \clipsimilarity(\video_{src}, \editedprompt)) + (\clipsimilarity(\video_{src}, \sourceprompt) - \clipsimilarity(\video_{edit}, \sourceprompt))$,
where $\sourceprompt$ is the source video description, $\editedprompt$ is the paired edited video description, $\video_{src}$ is the source video, $\video_{edit}$ is the edited video, and $\clipsimilarity(V, \prompt)$ is the Clip similarity between a video and a description.
Intuitively, we want $\video_{edit}$ to be more aligned with $\editedprompt$ than $\video_{src}$, and $\video_{src}$ to be more aligned with $\sourceprompt$ than $\video_{edit}$.

Since $\clipsimilarity(\video_{src}, \editedprompt)$ and $\clipsimilarity(\video_{src}, \sourceprompt)$ are identical for the samples within the group, they make no contribution to the gradient update due to the nature of group relative preference.
The instruction following score can then be formulated as:
\begin{align}
    \rewardscore_{\text{IF}} = \clipsimilarity(\video_{edit}, \editedprompt) - \clipsimilarity(\video_{edit}, \sourceprompt).
\end{align}

\paragraph{Source video preservation.}
A further key metric in video editing is that regions outside the edit should preserve high fidelity to the source video, and the edited content must also be coherent with the source video. 
We thus propose the following objective to quantify source video preservation reward score:
\begin{equation}
    \rewardscore_{\text{SP}} = \clipsimilarity(\video_{src}, \video_{edit}).
\end{equation}


\paragraph{Human preference alignment.}
To align video editing results with human preferences, we use Pickscore to comprehensively evaluate the alignment of the edited video with the instruction and its visual quality:
\begin{equation}
    \rewardscore_{\text{PS}} = \text{Pickscore}(\video_{edit}, \editedprompt).
\end{equation}

Overall, we use a combination of the aforementioned reward scores as the optimization objective to improve the model's capabilities of instruction following, source video preservation, and the human preference alignment.
\begin{equation}
    \rewardscore = \rewardweight_{IF}\rewardscore_{\text{IF}} + \rewardweight_{SP}\rewardscore_{\text{SP}} + \rewardweight_{PS}\rewardscore_{\text{PS}},
\end{equation}
where $\rewardweight_{IF}$, $\rewardweight_{SP}$, and $\rewardweight_{PS}$ are reward balancing weights.
Instead of full fine-tuning, we opt to train a LoRA~\cite{hu2022lora} on the DiT backbone during the Edit-GRPO stage for efficiency.
Please refer to the supplementary materials for the detailed LoRA design.

\subsection{Data Preparation}
\label{subsec:data_pipeline}

The field currently lacks a publicly available, large-scale, high-quality dataset of paired video editing examples. Existing datasets~\cite{zi2025se, bai2025scaling}, which often rely on inpainting or first-frame transfer guided by dense control signals (\eg, depth and edges), frequently suffer from visual artifacts, background inconsistencies, and inaccurate instructional alignment. To address this gap, we construct a large-scale video editing dataset comprising 1.5 million pairs, covering diverse local editing types and corresponding instructions.

Our data generation process begins with a pretrained text-to-video foundation model. We augment the MMDiT~\cite{esser2024scaling} architecture with an additional branch for video input, which is then fine-tuned on a meticulously designed dataset. This fine-tuned model is subsequently used to synthesize the video pairs. Empirically, we find that adding this dedicated branch significantly outperforms the typically-used ControlNet~\cite{zhang2023adding}.
We build our dataset following the steps described below, with more details available in the supplementary materials.

\input{figures/fig_VIEBench_qualitative_comparison}

\begin{description}[leftmargin=0pt, labelindent=0pt, parsep=0pt]
\item[Object replacement.]
We first extract the names of main objects from a detailed video caption. Then, we use Grounding-DINO~\cite{liu2023grounding} to detect the object on the first frame and employ SAM 2~\cite{ravi2024sam} to track its mask throughout the video. These masked video pairs are used to train a local inpainting model. Subsequently, we use an LLM to analyze the original video caption, rewrite it to reflect a new object, and use this new caption to inpaint the masked video, thereby synthesizing the editing pair.

\item[Object addition and removal.]
We introduce random masks onto a video and train an inpainting model to fill these regions. To synthesize an editing pair, we alter the original video's caption to remove the masked object and use this new caption for video inpainting. We also add the name of the masked object to the negative prompt to prevent a similar object from being inpainted in the masked region.
\item[Global content editing.]
We extract dense scribbles from the source video and train a model to synthesize new videos conditioned on this structural guidance and a modified caption. These modified captions are automatically generated by an LLM, which alters the original description (\eg, changing the background setting from ``day'' to ``night'').
\item[Data filtering.]
The synthesis process described above is inherently prone to artifacts.
Instead of relying on unreliable filtering, we employ a two-stage quality control strategy. First, we use a VLM to rewrite the instructions for the synthesized pairs. Second, we task the VLM with labeling whether the synthesized video is free from artifacts. We primarily use the synthesized video as the source and the real video as the target. We randomly choose a synthesized video as the target only when it is labeled as artifact-free. This setup ensures the editing model is not trained on degraded videos. After testing different VLMs, we select Gemini 2.5 Pro to write the detailed editing instructions.
\item[Image editing pairs.]
In addition, we collect 5.4 million image editing pairs from publicly available datasets~\cite{wang2025gpt, ge2024seed}. These image pairs are used to jointly train our editing model by treating each image as a single-frame video. We find that incorporating image data substantially enhances the diversity of editing types that the model can handle.
\end{description}

%% file: figures/fig_grpo.tex
\begin{figure}
\centering
\includegraphics[width=0.98\linewidth]{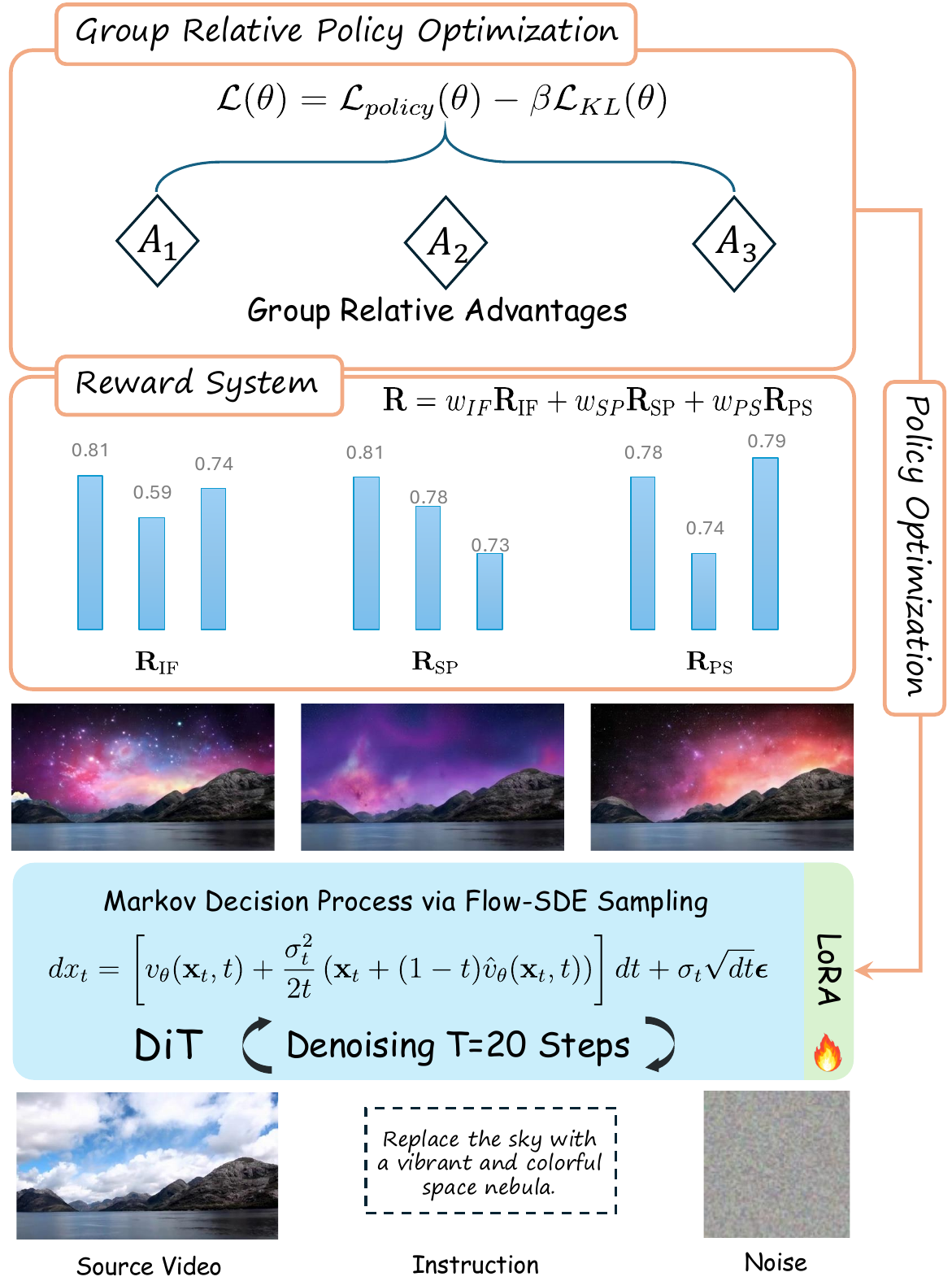} 
\caption{Overall pipeline of Edit-GRPO. 
We inject stochasticity via Flow-SDE~\cite{liu2025flowgrpo} to generate diverse samples, score them with our reward system, and compute a GRPO loss from the resulting relative advantages to update the model. 
For efficiency, we optimize a LoRA instead of full fine-tuning.
}
\label{fig:pipeline_grpo}
\end{figure}

%% file: figures/fig_VIEBench_qualitative_comparison.tex
\begin{figure*}
\centering
\includegraphics[width=0.98\textwidth]{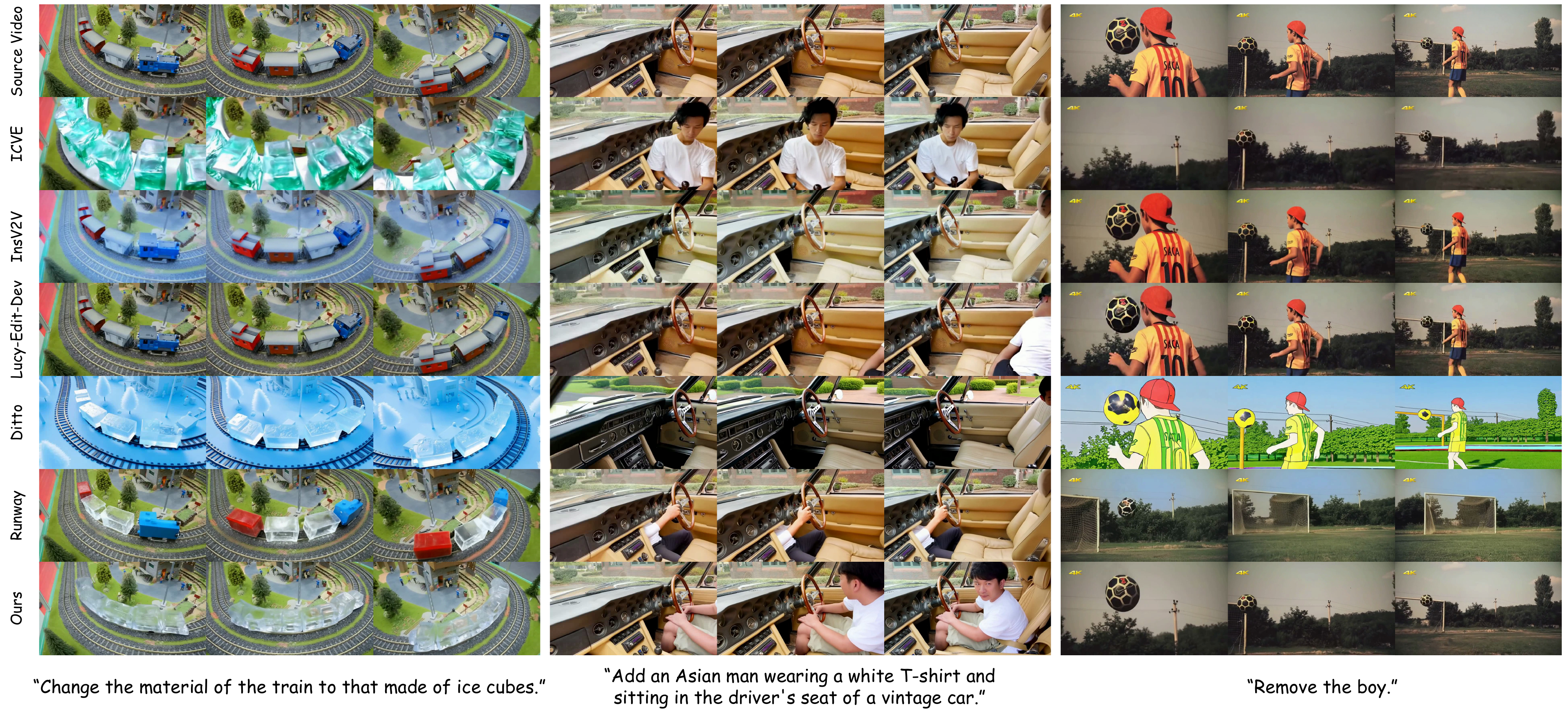}
\caption{Qualitative comparison of the instruction-based video editing on the VIE-Bench~\cite{mou2025instructx} dataset. The editing instruction corresponding to each group of results is shown at the bottom.
}
\label{fig:VIEBench_qualitative_comparison}
\end{figure*}

%% file: sec/6_experiment.tex
\section{Experiments}
\label{sec:experiments}

\input{tables/tab_viebench_comparison}

\subsection{Implementation Details}
\label{subsec:implementation details}
We use the pretrained \textit{HunyuanVideo-T2V-13B} as the DiT backbone and finetune it on our collected paired training data for $12,000$ steps.
We set the learning rate to $2 \times 10^{-5}$, with a global batch size of $128$.
To balance the mixing of image and video data, we sample video data with a probability of $0.4$ and image data with a probability of $0.6$.
Please refer to the supplementary materials for details.

\subsection{Experimental Setup}
\label{subsec:experiment_setup}

\paragraph{Benchmark and baselines.} 
To rigorously evaluate our model's instruction-based video editing capabilities, we utilize the VIE-Bench benchmark~\cite{mou2025instructx}, which comprises 140 high-quality instances across diverse instruction-based editing categories, covering both reference-free and reference-based edits.
We compare \modelname with five state-of-the-art instruction-based video editing methods, including four open-sourced ones, ICVE~\cite{liao2025iclunpaired}, Lucy-Edit-Dev~\cite{decart2025lucyedit}, Ditto~\cite{Bai2025Ditto}, and InsV2V~\cite{cheng2024insv2v}, and one commercial model, Runway Gen-4 Aleph ~\cite{runway2025aleph}.

\paragraph{Metrics.}
Since standard metrics like CLIP-based text-video alignment and PickScore~\cite{kirstain2023pick} are tailored for target-prompt-based editing evaluation, they are unable and unsuitable to evaluate the performance of instruction adherence.
Therefore, following prior works~\cite{mou2025instructx, bai2025scaling, ju2025editverse}, we employ a VLM evaluator(powered by Gemini-2.5-pro) to assign scores $(0-10)$ across four critical dimensions: Instruction Following (instruction adherence), Source Preservation (consistency with the source video), Editing Quality (overall video aesthetics), and Subject Similarity (for reference-based edits).
As an aiding metric, we also use the VBench~\cite{huang2024vbench} evaluation suite to provide quantitative measures of general video quality attributes.
We evaluate background consistency and subject consistency by calculating CLIP~\cite{radford2021clip} and DINO~\cite{caron2021emerging} feature similarity across frames respectively.
Following EditVerse~\cite{ju2025editverse}, we evaluate the frame-text CLIP~\cite{radford2021clip} feature alignment and video-text ViCLIP~\cite{wang2023internvid} feature alignment to reflect both semantics and style consistency.
We also evaluate the edited video quality from human preferences perspective by calculating frame-wise PickScore~\cite{kirstain2023pick}.

\subsection{Qualitative Comparisons}
\label{subsec:qualitative_comparison}
We conduct qualitative comparisons from two perspectives: instruction-only video editing and reference-based video editing, to provide a comprehensive evaluation.
Figure~\ref{fig:VIEBench_qualitative_comparison} presents instruction-only video editing results.
We select three representative task examples: local material change (``Change the material of the train to ice cubes.''), subject addition (``Add an Asian man wearing a white T-shirt and sitting in the driver's seat of a vintage car.''), and subject removal (``Remove the boy.'').
In the subject addition example, only our model correctly captures the complex semantics of ``sitting in the driver's seat.''
In the material change example, Runway fails to properly interpret the instruction, resulting in incomplete and inaccurate edits, modifying only part of the train.
In the removal example, only our model successfully preserves the soccer ball, removes the boy, and maintains background consistency with the source video.
These comparisons demonstrate that our model consistently outperforms the baselines, exhibiting superior editing capacity and generalization. 
We attribute this to our VLM instructor, which supplies the editing backbone with sufficiently rich context, enabling the model to comprehend complex editing semantics and to interpret how these semantics should be applied to the source video.

Since existing open-source instruction-based video editing models do not support reference-image control, we compare our model only with Runway Gen-4 Aleph.
As shown in Figure~\ref{fig:teaser}, although Runway responds to the instruction, it fails to accurately align with the reference image.
The hat and front label of the teddy bear show noticeable deviations from the reference in the edited results.
In contrast, our model correctly understands the intricate relationships among the instruction, the reference image, and the source video, and produces precise editing outcomes. 
This is enabled by encoding the reference image features into the VLM instructor module, 
providing ample context to accurately guide our DiT backbone.
Please refer to supplementary materials for more results on reference-image controlled video editing.

\input{tables/tab_ablation}
\input{figures/fig_userstudy}

\subsection{Quantitative Comparisons}
\label{subsec:quantitative_comparison}
As shown in Table~\ref{tab:VIEBench}, the quantitative metrics show consistent results with the qualitative comparisons. 
Our model achieves superior performance over all open-source baselines across all VLM-based evaluations and delivers results on par with the commercial model Runway, demonstrating our model's strong and robust video editing capability.
Since human judgments are more indicative of real-world effectiveness in video editing, we conducted a comprehensive user study with $14$ domain experts across three dimensions, following a setup similar to that used in VLM evaluations: (1) Instruction Following: whether the edit precisely follow the editing instruction; (2) Source Preservation: whether the edit maintains coherence with the original video content; (3) Editing Quality: whether the edited video is visually seamless, natural-looking, and pleasing.
We randomly sample $30$ instruction-only and $10$ reference-image-controlled video editing data.
Figure~\ref{fig:userstudy} quantifies human preference for comparison of generated video from ours versus baselines. 
Each comparison is labeled as one of three outcomes: $G$ (ours preferred), $S$ (no clear preference / indistinguishable), or $B$ (baseline preferred). 
Our method is preferred across all three criteria against five baselines.

\subsection{Ablation Study}
\label{subsec:ablation_study}
We conduct comprehensive ablations on multiple design choices of our model in Table~\ref{tab:ablation}. 
The introduction of the VLM instructor in Section~\ref{subsec:incontext_multimodality_understanding} significantly improves performance across all evaluation metrics. 
It delivers decisive gains in correctly and precisely following editing instructions, 
demonstrating that feeding visual information into the VLM instructor 
yields context with rich semantics, which then better guides the DiT.
The masked loss in Eq.~\ref{eq:maskloss} introduces an effective spatial inductive bias, 
enhancing overall editing capability, 
speeding up convergence, and producing more accurate responses within the edited regions.
The strategy of mixing image-editing data during training leverages the larger scale and broader edit-category coverage of image data to significantly improve both editing ability and visual quality. 
It also enables the emergence of many complex video editing types for which paired video data are typically difficult to obtain.
Our full model, with Edit-GRPO, exhibits comprehensive gains in overall editing performance. 
With the reward system designed in Section~\ref{subsec:posttraining_grpo}, the model achieves substantial improvements in instruction following, maintaining consistency with the source video. 
The higher PickScore~\cite{kirstain2023pick} indicates that the model trained with Edit-GRPO aligns better with human preferences.



%% file: tables/tab_viebench_comparison.tex
\begin{table*}
\centering
\resizebox{0.98\linewidth}{!}{
\begin{tabular}{l|l|l|ccccc|ccccc}
\toprule
\multirow{3}{*}{\textbf{Category}} &
\multirow{3}{*}{\textbf{Task}} &
\multirow{3}{*}{\textbf{Method}} &
\multicolumn{5}{c|}{\textbf{VLM Evaluation Score}} &
\multicolumn{5}{c}{\textbf{Video Quality}} \\
\cline{4-8} \cline{9-13}
& & &
\makecell{Instruction\\Following} &
\makecell{Source\\Preservation} &
\makecell{Editing\\Quality} &
\makecell{Subject\\Similarity} &
\makecell{Avg.} &
\makecell{CLIP\\Consistency} &
\makecell{DINO\\Consistency} &
\makecell{Pick-\\Score} &
\makecell{Frame-Text\\Alignment} &
\makecell{Video-Text\\Alignment} \\
\midrule
\multirow{24}{*}{\cellcolor{white} Instruction-only} &
\multirow{6}{*}{\cellcolor{white} Add}
& \textcolor{mygray}{Runway} & \textcolor{mygray}{9.44} & \textcolor{mygray}{9.16} & \textcolor{mygray}{6.72} & \textcolor{mygray}{-} & \textcolor{mygray}{8.44} & \textcolor{mygray}{97.95} & \textcolor{mygray}{98.39} & \textcolor{mygray}{21.40} & \textcolor{mygray}{28.40} & \textcolor{mygray}{26.30} \\
& & ICVE       &  \underline{8.28} & \underline{8.32} & \underline{5.08} & - & \underline{7.22} & \underline{97.99} & \underline{98.54} & \underline{20.53} & \underline{27.34} & \underline{25.60} \\
&  & InsV2V       &  5.92 & 5.96 & 2.80 & - & 4.89 & 96.80 & 97.32 & 19.88 & 24.53 & 21.90 \\
&  & Lucy-Edit-Dev   & 6.96  & 5.40 & 3.04 & - & 5.13 & 97.86 & 98.40 & 20.17 & 25.42 & 23.38 \\
&  & Ditto         & 5.92  & 7.44 & 3.72 & - & 5.69 & \textbf{98.26} & \textbf{98.91} & 20.51 & 26.30 & 24.29 \\
&  &  \textbf{Ours}   & \textbf{9.72} & \textbf{9.80} & \textbf{7.08} & - & \textbf{8.86} & 97.65 & 97.70 & \textbf{20.90} & \textbf{27.68} & \textbf{26.32} \\
\cmidrule{2-13}
& \multirow{6}{*}{\cellcolor{white} \makecell{Replace}}
& \textcolor{mygray}{Runway} & \textcolor{mygray}{9.91} & \textcolor{mygray}{9.44} & \textcolor{mygray}{7.79} & \textcolor{mygray}{-} & \textcolor{mygray}{9.04} & \textcolor{mygray}{96.38} & \textcolor{mygray}{96.23} & \textcolor{mygray}{21.18} & \textcolor{mygray}{26.64} & \textcolor{mygray}{26.24} \\
&  & ICVE   & 8.20  & \underline{8.38} & \underline{4.50} & - & \underline{7.02} & 95.95 & 95.18 & \underline{20.48} & \underline{26.06} & \textbf{25.96} \\
&  & InsV2V       & 6.17  & 6.29 & 3.41 & - & 5.29 & 95.14 & 93.78 & 20.46 & 25.15 & 24.62 \\
&  & Lucy-Edit-Dev         &  \underline{8.64} & 6.38 & 3.23 & - & 6.08 & 95.75 & 94.84 & 20.39 & 24.54 & 24.39 \\
&  & Ditto         &  3.70 & 5.35 & 3.61 & - & 4.22 & \textbf{96.78} & \textbf{96.61} & 20.17 & 24.26 & 22.13 \\
&  &  \textbf{Ours}         & \textbf{9.61} & \textbf{9.61} & \textbf{7.35} & - & \textbf{8.86} & \underline{95.97} & \underline{95.57} & \textbf{21.03} & \textbf{26.45} & \underline{25.82} \\
\cmidrule{2-13}
& \multirow{6}{*}{\cellcolor{white} \makecell{Remove}}
& \textcolor{mygray}{Runway} & \textcolor{mygray}{9.86} & \textcolor{mygray}{10.0} & \textcolor{mygray}{9.51} & \textcolor{mygray}{-} & \textcolor{mygray}{9.79} & \textcolor{mygray}{97.62} & \textcolor{mygray}{98.00} & \textcolor{mygray}{20.50} & \textcolor{mygray}{24.35} & \textcolor{mygray}{24.82} \\
&  & ICVE     &  \underline{7.24} & \underline{8.68} & \underline{5.20} & - & \underline{7.04} & \textbf{97.39} & \textbf{97.40} & \underline{19.98} & \underline{22.95} & \underline{23.30} \\
&  & InsV2V         &  5.34 & 6.72 & 2.03 & - & 4.70 & 95.97 & 95.70 & 19.32 & 21.56 & 18.76 \\
&  & Lucy-Edit-Dev           & 6.65  & 5.00 & 3.34 & - & 5.00 & 96.58 & 96.12 & 19.37 & 21.10 & 18.43 \\
&  & Ditto           & 4.27 & 4.27 & 3.72 & - & 4.09 & \underline{97.18} & \underline{96.99} & 19.15 & 20.44 & 15.63 \\
&  &  \textbf{Ours}           & \textbf{9.37} & \textbf{10.0} & \textbf{8.96} & - & \textbf{9.44} & 97.17 & 96.94 & \textbf{20.45} & \textbf{24.36} & \textbf{24.91} \\
\cmidrule{2-13}
& \multirow{6}{*}{\cellcolor{white} Hybrid Edit}
  & \textcolor{mygray}{Runway} & \textcolor{mygray}{8.77} & \textcolor{mygray}{8.00} & \textcolor{mygray}{6.77} & \textcolor{mygray}{-} & \textcolor{mygray}{7.85} & \textcolor{mygray}{95.88} & \textcolor{mygray}{95.85} & \textcolor{mygray}{21.50} & \textcolor{mygray}{29.57} & \textcolor{mygray}{30.37} \\
&  & ICVE     &  5.55 & \underline{5.66} & \underline{3.55} & - & \underline{4.92} & \underline{96.01}  & \underline{96.60} & 20.17 & 26.70 & 25.38 \\
&  & InsV2V         & 4.55 & 3.44 & 2.55 & - & 3.51 & 94.66 & 95.27 & 20.41 & \underline{27.76} & 24.93 \\
&  & Lucy-Edit-Dev   &  \underline{6.66} & 4.11 & 3.33 & - & 4.70 & 95.50 & 95.61 & \underline{20.46} & 26.31 & 25.68 \\
&  & Ditto       & 2.00  & 3.11 & 1.66 & - & 2.25 & \textbf{96.76} & \textbf{96.88} & 20.25 & 26.49 & \underline{25.88} \\
&  &  \textbf{Ours}         & \textbf{6.88}  & \textbf{6.66} & \textbf{4.11} & - & \textbf{5.88} & 94.99 & 95.67 & \textbf{20.82} & \textbf{27.87} & \textbf{28.72} \\
 \midrule
 \multirow{4}{*}{\cellcolor{white} Reference-based} &
 \multirow{2}{*}{\cellcolor{white} \makecell{Add}}
 & \textcolor{mygray}{Runway} & \textcolor{mygray}{10.0} & \textcolor{mygray}{10.0} & \textcolor{mygray}{7.20} & \textcolor{mygray}{8.60} & \textcolor{mygray}{8.95} & \textcolor{mygray}{96.22} & \textcolor{mygray}{96.43} & \textcolor{mygray}{20.12} & \textcolor{mygray}{28.36} & \textcolor{mygray}{27.62} \\
&  &  \textbf{Ours}      &  9.90 & \textbf{10.0} & 7.00 & \textbf{8.96} & \textbf{8.96} & 95.92 & \textbf{96.52} & 19.97 & 28.11 & 27.23 \\
\cmidrule{2-13}
& \multirow{2}{*}{\cellcolor{white} \makecell{Replace}}
&  \textcolor{mygray}{Runway} & \textcolor{mygray}{9.44} & \textcolor{mygray}{8.22} & \textcolor{mygray}{5.77} & \textcolor{mygray}{8.88} & \textcolor{mygray}{8.07} & \textcolor{mygray}{97.06} & \textcolor{mygray}{97.66} & \textcolor{mygray}{21.37} & \textcolor{mygray}{28.49} & \textcolor{mygray}{29.67} \\
&  &  \textbf{Ours}      & \textbf{9.77}  & \textbf{9.88} & \textbf{5.88}  & \textbf{9.44} & \textbf{8.74} & 97.01 & 97.60 & 21.20 & \textbf{28.71} & \textbf{30.19} \\
\bottomrule
\end{tabular}%
}
\caption{Quantitative comparison results on the VIE-Bench~\cite{mou2025instructx} dataset.
The best and second-best results of the open-sourced methods are highlighted in \textbf{bold} and \underline{underlined}, respectively.
As reference-based video editing is not yet supported by the open-source models, we provide our results against the commercial model Runway Gen-4 Aleph ~\cite{runway2025aleph}.
}
\label{tab:VIEBench}
\end{table*}

%% file: tables/tab_ablation.tex
\begin{table*}
\centering
\resizebox{0.98\linewidth}{!}{
\begin{tabular}{l|l|cccc|ccccc}
\toprule
\multirow{3}{*}{\textbf{Task}} &
\multirow{3}{*}{\textbf{Method}} &
\multicolumn{4}{c|}{\textbf{VLM Evaluation Score}} &
\multicolumn{5}{c}{\textbf{Video Quality}} \\
\cline{3-6} \cline{7-11}
& &
\makecell{Instruction\\Following} &
\makecell{Source\\Preservation} &
\makecell{Editing\\Quality} &
\makecell{Avg.} &
\makecell{CLIP\\Consistency} &
\makecell{DINO\\Consistency} &
\makecell{Pick-\\Score} &
\makecell{Frame-Text\\Alignment} &
\makecell{Video-Text\\Alignment} \\
\midrule
\multirow{5}{*}{\cellcolor{white} Add}
 & Vanilla        & 6.44  & 4.56 & 2.72 & 4.57 & 97.49 & \textbf{98.85} & 19.65 & 23.38 & 20.02 \\
 & + V      & 8.80  & 7.28 & 4.52  & 6.86 & \textbf{98.27} & \underline{98.63} & 20.53 & 26.90 & 25.28 \\
 & + V + M       & \underline{8.84} & \underline{8.68} & \underline{6.92} & \underline{8.14} & 97.33 & 97.97 & 20.60 & 27.38 & 26.13 \\
 & + V + M + I   & 8.76 & 8.32 & 6.68 & 7.91 & 97.56 & 98.22 & \textbf{21.03} & \textbf{27.90} & \underline{26.20} \\
 & + V + M + I + E  & \textbf{9.72} & \textbf{9.80} & \textbf{7.08} & \textbf{8.86} & \underline{97.65} & 97.70 & \underline{20.90} & \underline{27.68} & \textbf{26.32} \\
 \midrule
 \multirow{5}{*}{\cellcolor{white} \makecell{Replace}}
 & Vanilla        &  4.91  & 3.70 & 3.47 & 4.02 & 95.63 & 95.27 & 19.87 & 21.33 & 18.90 \\
 & + V      & 8.23  & 6.23 & 5.05 & 6.50 & 96.17 & 95.07 & 20.46 & 24.29 & 24.06 \\
 & + V + M       & 7.70 & 6.91 & 4.94  & 6.51 & \underline{95.79} & 95.10 & 20.32 & 24.47 & 23.86 \\
 & + V + M + I    &  \textbf{9.64} & \underline{9.50} & \underline{7.32} & \underline{8.82} & 95.73 & \underline{95.57} & \underline{20.98} & \underline{26.44} & \textbf{26.25} \\
 & + V + M + I + E  & \underline{9.61} & \textbf{9.61} & \textbf{7.35} & \textbf{8.86} & \textbf{95.97} & \textbf{95.57} & \textbf{21.03} & \textbf{26.45} & \underline{25.82} \\
\midrule
\multirow{5}{*}{\cellcolor{white} Remove}
 & Vanilla        & 2.13  & 3.75 & 1.96 & 2.62 & 96.44 & 96.00 & 18.83 & 20.49 & 16.66 \\
 & + V      &  8.00 & 8.51 & 6.44  & 7.65 & \textbf{97.82} & 96.77 & 20.17 & 23.31 & 23.38 \\
 & + V + M       & 8.72 & 9.06 &  7.00 & 8.26 & \underline{97.35} & \textbf{97.80} & 20.19 & 24.19 & 24.44 \\
 & + V + M + I    &   \underline{9.20}  & \underline{9.82} & \underline{8.48}  & \underline{9.17} & 97.30 & \underline{97.57} & \underline{20.36}  & \textbf{24.44} & \textbf{25.25} \\
 & + V + M + I + E  & \textbf{9.37} & \textbf{10.0} & \textbf{8.96} & \textbf{9.44} & 97.17 & 96.94 & \textbf{20.45} & \underline{24.36} & \underline{24.91} \\
\bottomrule
\end{tabular}%
}
\caption{Ablation study results on VIE-Bench.
V: VLM instructor; M: masked loss; I: Mixing image data; E: Edit-GRPO.
The last row for each task corresponds to our full method.
The best and second-best results are highlighted in \textbf{bold} and \underline{underlined}, respectively.
}
\label{tab:ablation}
\end{table*}

%% file: figures/fig_userstudy.tex
\begin{figure*}
\centering
\includegraphics[width=0.98\textwidth]{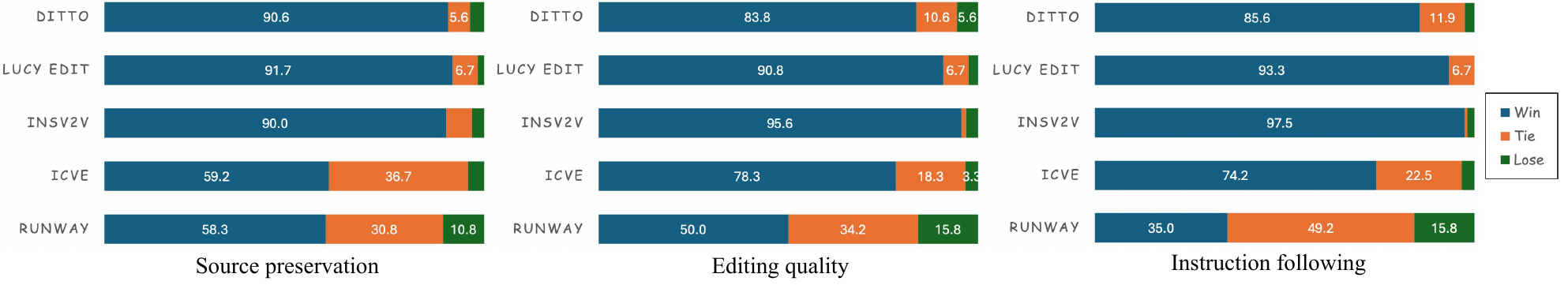} 
\caption{User study results. We conduct 1-to-1 paired comparisons. Win indicates users prefer ours better than baseline, vice versa. Tie indicates there is no significant difference. Numbers are reported in percentage. 
}
\vspace{-0.02in}
\label{fig:userstudy}
\end{figure*}

%% file: sec/8_conclusion.tex
\section{Conclusion}
\label{sec:conclusion}
In this work, we present \modelname, a scalable framework for instruction-based video editing that outperforms existing open-source baselines, achieving state-of-the-art results in terms of instruction following, edit quality, and generalization.
We introduce a VLM-based instructor that encodes both textual instructions and visual inputs into grounded multimodal tokens, substantially improving semantic disambiguation and edit controllability while preserving coherence with the source video.
We also propose Edit-GRPO, which, for the first time, adapts Group Relative Policy Optimization to video editing and yields consistent performance gains.

Future work includes expanding the range of supported edit types toward richer and more compositional behaviors, enhancing practical usability by targeting interactive editing, as well as extending to arbitrary resolution and duration. 
Another promising direction is developing a more general and unified reward model tailored for video editing.
Limitations and failure cases are discussed in the supplementary materials.

%% file: appendix/preliminary.tex
\section{Preliminaries}
\label{sec:preliminaries}

\subsection{DiT-Based Video Generation and Editing}
\label{sec:preliminary:dit}

Diffusion Transformers (DiTs) have become a powerful backbone for modern video generation.
They model a sequence of latent video tokens through a transformer-based denoising process conditioned on multimodal inputs such as text or reference frames.
DiTs are commonly trained with the \textit{Flow Matching} objective~\cite{lipman2022flow}, which learns a velocity field $\mathbf{v}_\theta(\mathbf{x}, t)$ that transports a simple prior $p_0(\mathbf{x})$ (\eg, Gaussian noise) toward a target data distribution $p_1(\mathbf{x})$ via a probability flow Ordinary Differential Equation (ODE):
\begin{equation}
\frac{d\mathbf{x}_t}{dt} = \mathbf{v}_\theta(\mathbf{x}_t, t), \quad \text{with} \quad \mathbf{x}_0 \sim p_0(\mathbf{x}).
\end{equation}
In the commonly used \textit{Rectified Flow} formulation~\cite{liu2022flow}, the trajectory is defined as a linear interpolation between noise and data samples:
\begin{equation}
\mathbf{x}_t = (1-t)\mathbf{x}_0 + t\mathbf{x}_1, \quad \mathbf{x}_1 \sim p_1(\mathbf{x}),
\end{equation}
and the model is trained to regress the target velocity field
$\mathbf{u}_t(\mathbf{x}_t) = \frac{d\mathbf{x}_t}{dt} = \mathbf{x}_1 - \mathbf{x}_0$
via the loss:
\begin{equation}
\mathcal{L}_{\mathrm{FM}} =
\mathbb{E}_{t, \mathbf{x}_0, \mathbf{x}_1}\!\left[
\|\mathbf{v}_\theta(\mathbf{x}_t, t) - (\mathbf{x}_1 - \mathbf{x}_0)\|_2^2
\right].
\label{eq:rectified_flow_loss}
\end{equation}
This objective learns a continuous flow that transports noisy latents toward clean samples, providing smoother temporal dynamics and faster convergence than conventional noise-prediction objectives.

For video editing, the DiT backbone is conditioned on the source video and an editing instruction, enabling it to synthesize instruction-aligned output based on the input video.


\subsection{Group Relative Policy Optimization (GRPO)}
\label{sec:preliminary:grpo}

Group Relative Policy Optimization (GRPO)~\cite{shao2024deepseekmath} is a recent reinforcement learning algorithm that extends Proximal Policy Optimization (PPO)~\cite{schulman2017ppo} for efficient post-training alignment.
For each input query $c$, a group of $G$ sampled trajectories $\{ \mathbf{x}_t^{i}\}_{i=1, \cdots, G}^{t=0, \cdots, T}$ are generated and evaluated by the task-specific reward criterion to get the corresponding reward signals $\{ R(\mathbf{x}_0^{i})\}_{i=0}^G$.
These reward signals are then transformed to relative advantages within the group by:
\begin{equation}
    \hat{A}_t^i = \frac{R(\mathbf{x}_0^{i}) - \text{mean}(\{R(\mathbf{x}_0^{i})\}_{i=1}^G)}{\text{std}(\{R(\mathbf{x}_0^{i})\}_{i=1}^G)}.
\end{equation}
This relative formulation reduces variance and stabilizes optimization, making GRPO effective for aligning large generative models with human or semantic preferences.

The policy model is then optimized by maximizing the following objective without requiring a separate critic network:
\begin{align}
    \mathcal{L}(\theta) &= \mathbb{E}_{c, \{\mathbf{x}^i\}_{i=1}^G \sim \pi_{\theta_{\text{old}}}(\cdot|c)}f(r, \hat{A}, \theta, \epsilon, \beta) \nonumber \\
    &= \frac{1}{G} \sum_{i=1}^G \frac{1}{T} \sum_{t=0}^{T-1}(\mathcal{L}_{policy}(\theta) - \beta \mathcal{L}_{KL}(\theta)),
\end{align}
where
\begin{align}
    \mathcal{L}_{policy}(\theta) &= \min(r_t^i(\theta)\hat{A}_t^i, \text{clip}(r_t^i(\theta), 1 - \epsilon, 1 + \epsilon)\hat{A}_t^i) \nonumber, \\
    \mathcal{L}_{KL}(\theta) &=D_{\text{KL}}(\pi_{\theta}||\pi_{\text{ref}}) \nonumber,\\
    r_t^i(\theta) &= \frac{p_{\theta}(\mathbf{x}_{t-1}^i \mid \mathbf{x}_t^i, \mathbf{c})}{p_{\theta_{\text{old}}}(\mathbf{x}_{t-1}^i \mid \mathbf{x}_t^i, \mathbf{c})}.
\end{align}

Since GRPO relies on stochastic sampling diverse trajectories $\{ \mathbf{x}_t^{i}\}_{i=1, \cdots, G}^{t=0, \cdots, T}$, Flow-GRPO~\cite{liu2025flowgrpo} introduce randomness into Flow Matching by converting the deterministic Flow-ODE into an equivalent Flow-SDE
\begin{equation}
    dx_t = \left[v_{\theta}(\mathbf{x}_t, t) + \frac{\sigma_t^2}{2t}\left(\mathbf{x}_t + (1-t)\hat{v}_{\theta}(\mathbf{x}_t, t)\right)\right]dt + \sigma_t\sqrt{dt}\boldsymbol{\epsilon}
    \label{eq:flowsde}
\end{equation}
where $\boldsymbol{\epsilon} \sim \mathcal{N}(0, I)$ is a newly sampled gaussian noise, $\sigma_t = \eta \sqrt{\frac{t}{1-t}}$ for Flow-GRPO.

Coefficients-Preserving Sampling~\cite{wang2025coefficients} investigates the problem of an excess of noise injected during Flow-SDE sampling in Eq.~\ref{eq:flowsde} and reformulates the sampling process to eliminate the noise artifacts by modifying $\sigma_t=\text{sin}(\frac{\eta\pi}{2})dt$.

%% file: appendix/implementation_details.tex
\section{Implementation Details}
\label{appendix:sec:Implementation_details}
During Edit-GRPO, We insert Low-Rank Adaptation (LoRA)~\cite{hu2022lora} modules into the self-attention and cross-attention layers of the DiT~\cite{kong2025hunyuanvideosystematicframeworklarge}.
We freeze the parameters of the model after supervised fine-tuning.
Only the LoRA parameters are updated, facilitating efficient optimization.
For the LoRA configuration, we set the low-rank dimension to $r=64$ and the scaling factor to $\alpha=128$. 
The adapter weights are initialized using a Gaussian distribution.

During inference, we use a classifier-free guidance scale of $2.0$, applying it only to instruction conditions. 
The inference timestep is set to $50$ for the balance of performance and inference speed.

%% file: appendix/data_strategy.tex
\section{Data Strategy}
\label{appendix:subsec:datastrategy}

\input{figures/fig_data_architecture}
\subsection{Detailed Architecture for Data Preparation}
Having briefly outlined the pipeline for constructing editing pairs in the main text, we now provide the detailed network architecture used for data synthesis. As illustrated in Figure~\ref{fig:data_architecture}, we integrate an additional branch into the MMDiT architecture~\cite{esser2024scaling} to model the interaction between the VAE embeddings of the conditions (e.g., masked video or scribbles) and the noisy latent. The condition tokens and the noisy latent tokens are concatenated, enabling full mutual attention across the entire combined sequence. Furthermore, we align the Rotary Positional Embeddings (RoPE) of the condition tokens with those of the noisy latent tokens. This design is based on the assumption that spatially and temporally corresponding positions in the condition inputs and the target videos are highly correlated and should thus exhibit the highest attention values.

To reduce computational complexity, we insert this additional branch block once every four blocks within the pretrained DiT. During training, we only update the parameters of this additional branch while keeping the pretrained DiT weights frozen to preserve the generative capabilities of the foundation model.

Empirically, we observe that the standard ControlNet architecture is highly sensitive to the Classifier-Free Guidance (CFG) scale. It often generates oversaturated samples at high CFG values, while suffering from poor instruction following at low CFG values. In contrast, our model remains free from saturation artifacts across a wide range of CFG values. We attribute this improvement to the bi-directional attention mechanism between the conditional input and the noisy latent enabled by our branch. Figure~\ref{fig:sup:training_data} visualizes representative training samples spanning addition, removal, replacement, and stylization tasks.
\input{figures/fig_sup_training_data}

\subsection{Edit-GRPO Data}
For the Edit-GRPO training stage, only the source videos $\sourcevideo$ are required. 
We curate a subset of high-quality videos and employ Gemini 2.5 Pro~\cite{gemini2025} to generate suitable editing instructions $\instruction$ for each source video. 
These prompts are specifically designed to encompass complex combinations of various editing types.
We leverage Gemini 2.5 Pro to caption the source video to get the source video prompt $\sourceprompt$. 
Subsequently, conditioned on $\sourceprompt$ and $\instruction$, Gemini 2.5 Pro generates the target video caption $\editedprompt$. 
This pipeline yields the triplet $(\sourcevideo, \sourceprompt, \editedprompt)$ required for training Edit-GRPO.

%% file: figures/fig_data_architecture.tex
\begin{figure}
\centering
\includegraphics[width=0.98\linewidth]{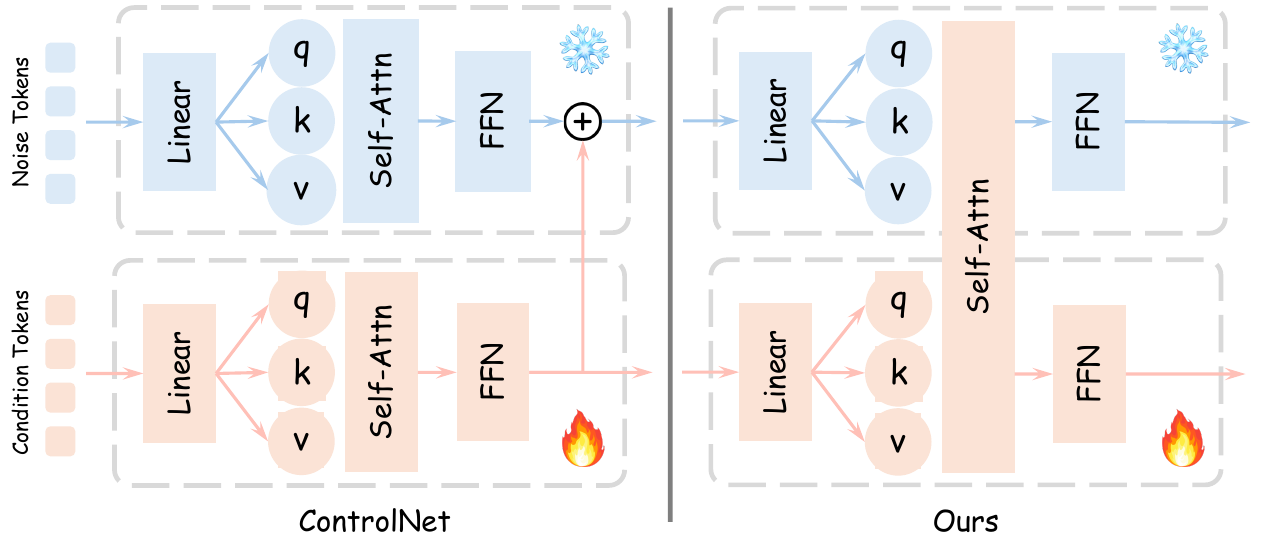} 
\caption{Network architecture for paired data synthesis. We modify the pretrained MMDiT backbone by inserting an additional control branch every four blocks. By concatenating the condition tokens with the noisy latent, the model performs full mutual attention across the sequence, ensuring robust structural alignment between the control signal and the generated video.
}
\label{fig:data_architecture}
\end{figure}

%% file: figures/fig_sup_training_data.tex
\begin{figure*}
\centering
\includegraphics[width=0.95\textwidth]{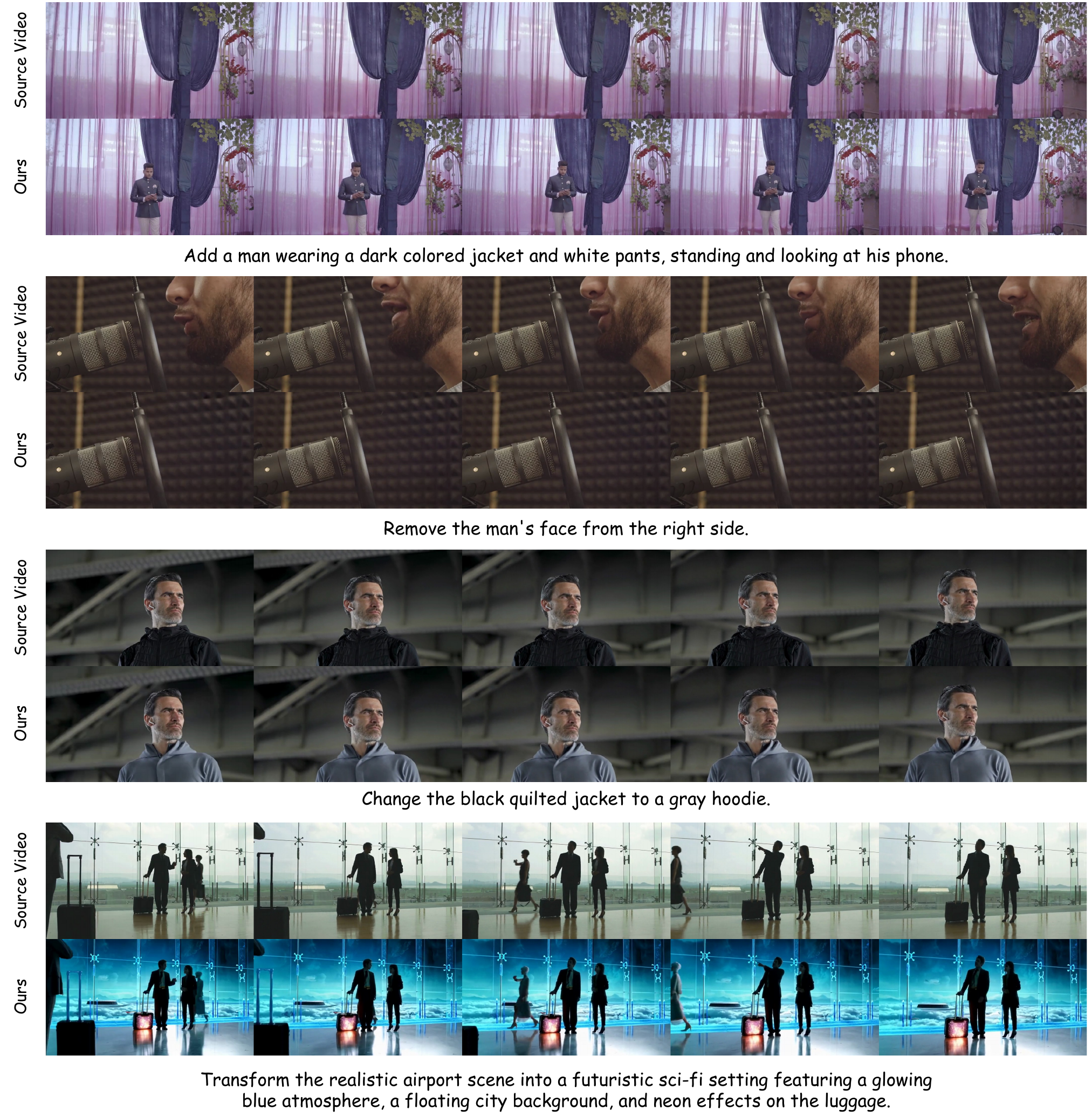} 
\caption{We visualize representative training samples spanning addition, removal, replacement, and stylization tasks, arranged from top to bottom. 
}
\label{fig:sup:training_data}
\end{figure*}

%% file: appendix/experiments.tex
\input{figures/fig_sup_comparison_ref}
\input{figures/fig_sup_failure}

\input{figures/fig_sup_ablation_grpo}

\input{figures/fig_sup_qualitative_0}
\input{figures/fig_sup_qualitative_1}
\input{figures/fig_sup_qualitative_2}

\section{Experiments}
\label{appendix:sec:experiments}

\subsection{Qualitative Comparisons}
\label{appendix:subsec:qualitative_ref}
Figure~\ref{fig:sup:qualitative_ref} presents qualitative comparisons of the reference-based video editing on the VIE-Bench~\cite{mou2025instructx}. 
Since existing open-source instruction-based video editing models do not support reference-image control, we compare our model only with Runway Gen-4 Aleph.
As shown in Figure~\ref{fig:sup:qualitative_ref}, although Runway responds to the instruction, it fails to accurately align with the reference image.
For example, the toy car and red backpack show noticeable deviations from the reference in the edited results.
In contrast, our model correctly understands the intricate relationships among the instruction, the reference image, and the source video, and produces precise editing outcomes.

\subsection{Ablation Study on Edit-GRPO}
\label{appendix:subsec:ablation_editgrpo}
Figure~\ref{fig:sup:ablation_grpo} presents qualitative comparisons on before and after applying Edit-GRPO.
Edit-GRPO yields a substantial qualitative improvement, consistent with quantitative ablation results in the main paper.
In challenging prompts such as ``Add a woman with long black hair Lying in the bucket of a green agricultural loader.'' and ``Replace the man into a gorilla.'', Edit-GRPO preserves the structural integrity and plausibility of the subjects, whereas removing it tends to produce distorted artifacts and collapsing results.
For instances like ``Add a dinosaur standing on the ground of gas station.'' and ``Add a big white kitten in the car trunk.'', Edit-GRPO makes the edited areas more natural and realistic, showing exceptional consistency with the source video.
Overall, the model equipped with EditGRPO demonstrates stronger instruction adherence, higher aesthetic fidelity, more consistent with the source video, and better alignment with human preference.

\subsection{Complex Tasks}

Figures~\ref{fig:sup:pick_data_0}, \ref{fig:sup:pick_data_1}, and \ref{fig:sup:pick_data_2} present more results of our method on complex instructions that are non-trivial and challenging to be synthesized by the data construction pipeline.
Our model is capable of performing complex hybrid instructions like ``Make the mountain in the background an active volcano erupting with smoke and lava, and add a little cat running by the couple.'', and non-rigid visual effects such as flames, smokes, fireworks, stylization, and watermark removal.
Notably, our model effectively addresses a challenging case in global background editing: ensuring physical coherence between the foreground subject and the edited new environment. 
This is vividly demonstrated in the example ``Change the background from the ocean to a vast, snowy mountain range.''
In this scenario, our model successfully synthesizes realistic shadows cast by the subject onto the snowy terrain, thereby achieving a high degree of photorealism and subject-background harmony.
Since these complex editing scenarios are difficult to synthesize through the data construction pipeline, they serve as a rigorous test of a model's generalization capabilities. We attribute our model's success to two key factors:
First, our VLM instructor possesses an exceptional capacity to interpret the intricate correlation between textual instructions and visual contexts.
Second, our strategy of mixing image editing data during training effectively transfers the robust generalization inherent in the image domain to the video editing domain.

\input{figures/fig_sup_vlmtemplate}

\subsection{VLM Evaluation}
We employ Gemini-2.5-Pro~\cite{gemini2025} as the VLM evaluator.
Figure~\ref{fig:sup:vlm_template} presents our VLM evaluation templates for the instruction-based video editing and reference-instruction-based video editing.
The VLM evaluator offers a scalable, human-aligned assessment. 
It is provided with the source frame, the edited frame, the instruction and an optional reference image as inputs, and evaluates the editing performance on a scale of $0$ (worst) to $10$ (best) across three four critical dimensions: Instruction Following (instruction adherence), Source Video Preservation (consistency with the source video), Editing Quality (overall video aesthetics), and Subject Similarity (for reference-based edits).

%% file: figures/fig_sup_comparison_ref.tex
\begin{figure*}
\centering
\includegraphics[width=\linewidth]{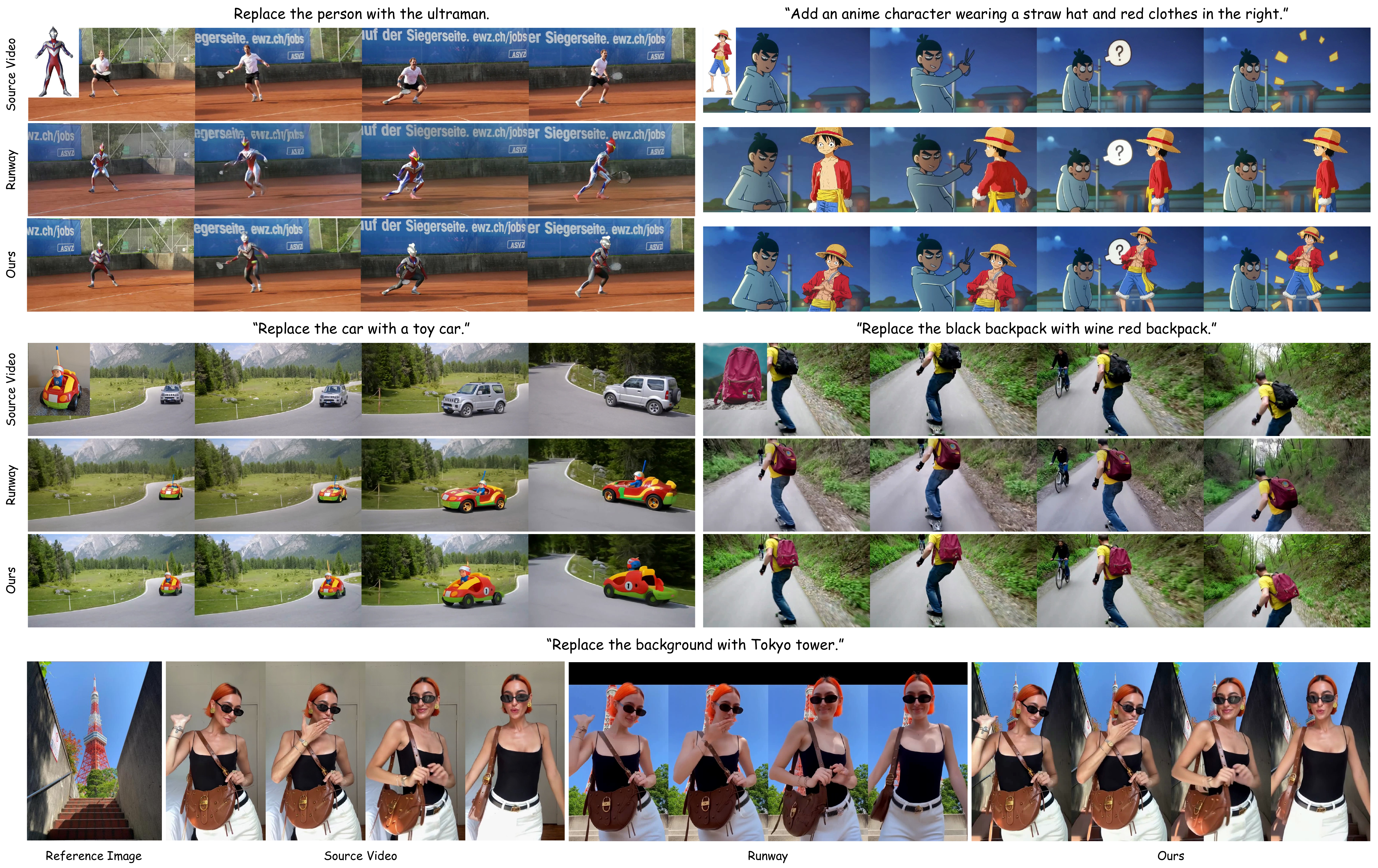} 
\caption{Qualitative comparison of the reference-based video editing on the VIE-Bench~\cite{mou2025instructx}. 
The editing instruction is shown at the top and the reference image is shown on the left for each group of results.
}
\label{fig:sup:qualitative_ref}
\end{figure*}

%% file: figures/fig_sup_failure.tex
\begin{figure*}
\centering
\includegraphics[width=\linewidth]{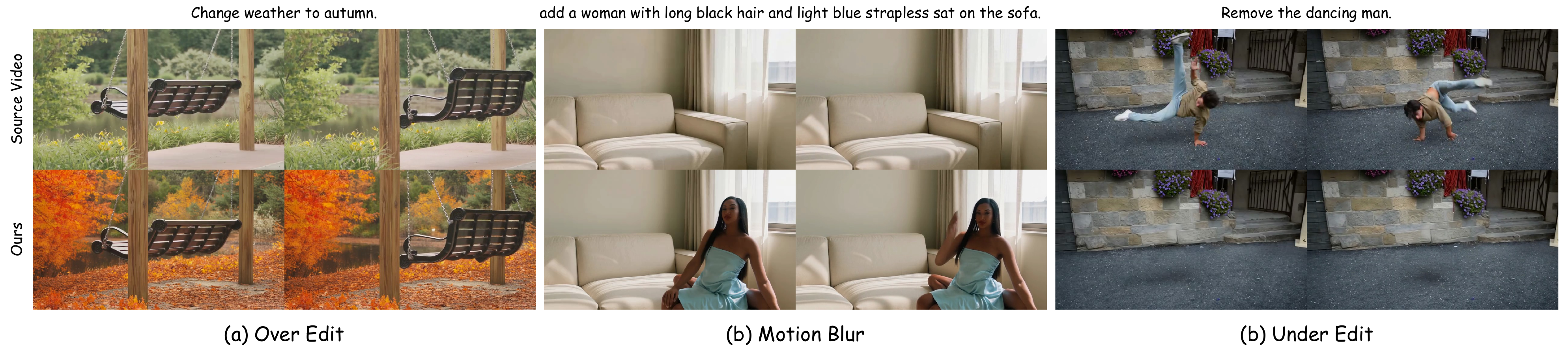} 
\caption{Failure cases. 
(a) Global transformations such as changing weather sometimes cause over-editing.
(a) Rapid motion might occasionally lead to blurry results, such as the woman's hand.
(c) Under-editing might be observed in removal tasks, where residual artifacts, such as cast shadows, remain.
}
\label{fig:sup:failure_cases}
\end{figure*}

%% file: figures/fig_sup_ablation_grpo.tex
\begin{figure*}
\centering
\includegraphics[width=0.98\textwidth]{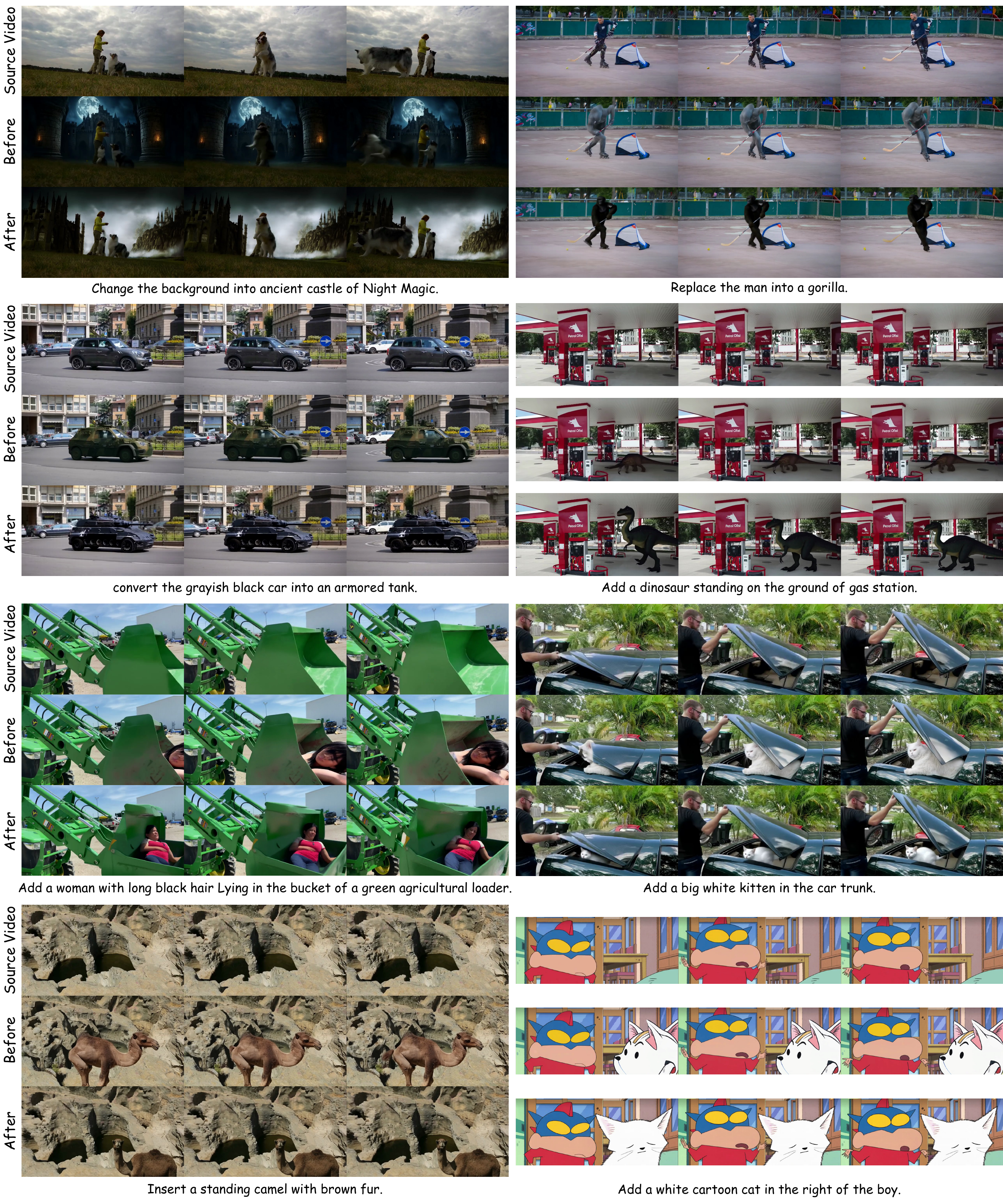} 
\caption{Ablation studies on Edit-GRPO. Before: without Edit-GRPO; After: with Edit-GRPO.
The editing instruction is shown at the bottom.
}
\label{fig:sup:ablation_grpo}
\end{figure*}

%% file: figures/fig_sup_qualitative_0.tex
\begin{figure*}
\centering
\includegraphics[width=0.95\textwidth]{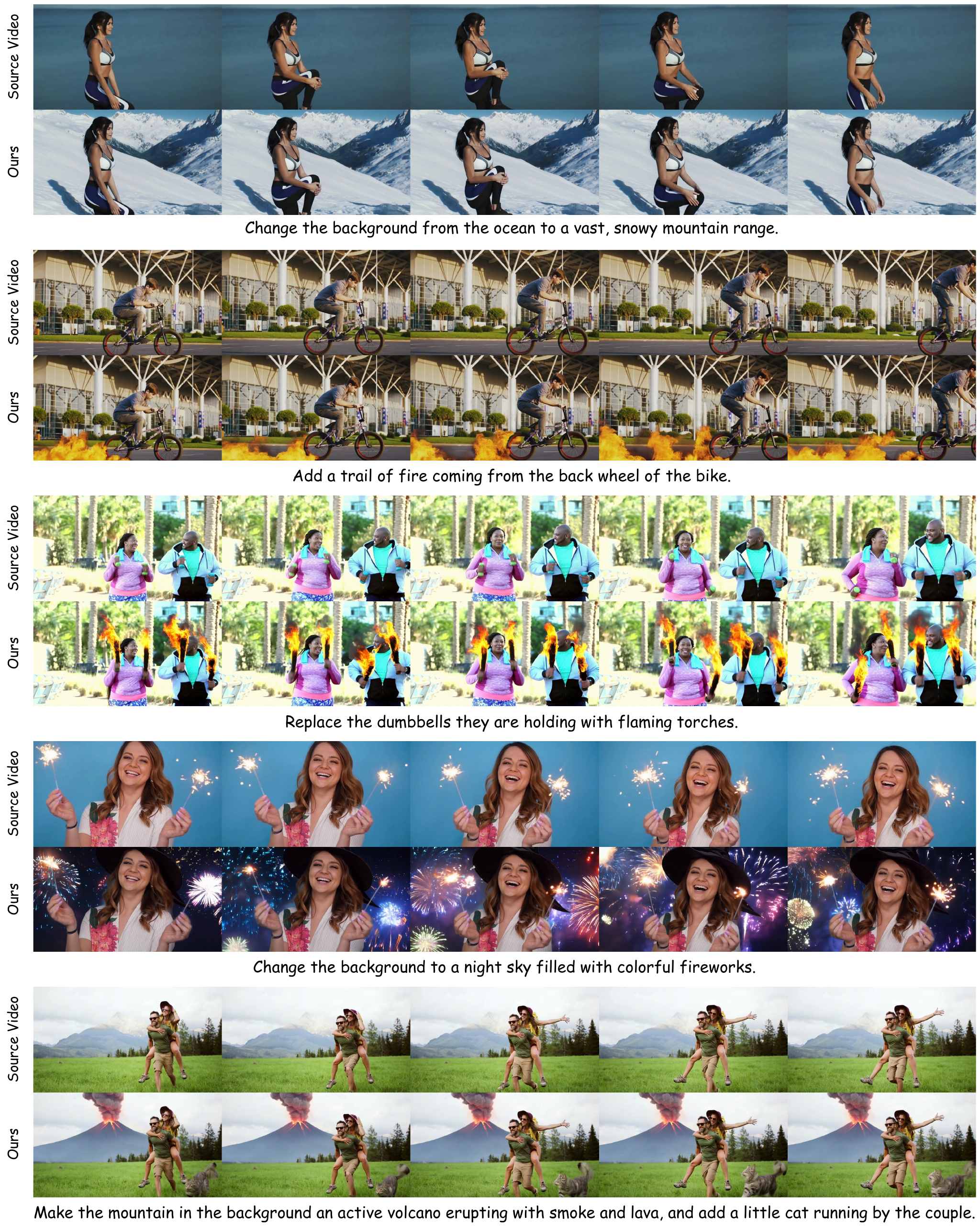} 
\caption{Qualitative results. We present more results of our method on complex instructions that are non-trivial and challenging to be synthesized by the data construction pipeline.
The editing instruction is shown at the bottom.
}
\label{fig:sup:pick_data_0}
\end{figure*}

%% file: figures/fig_sup_qualitative_1.tex
\begin{figure*}
\centering
\includegraphics[width=0.96\textwidth]{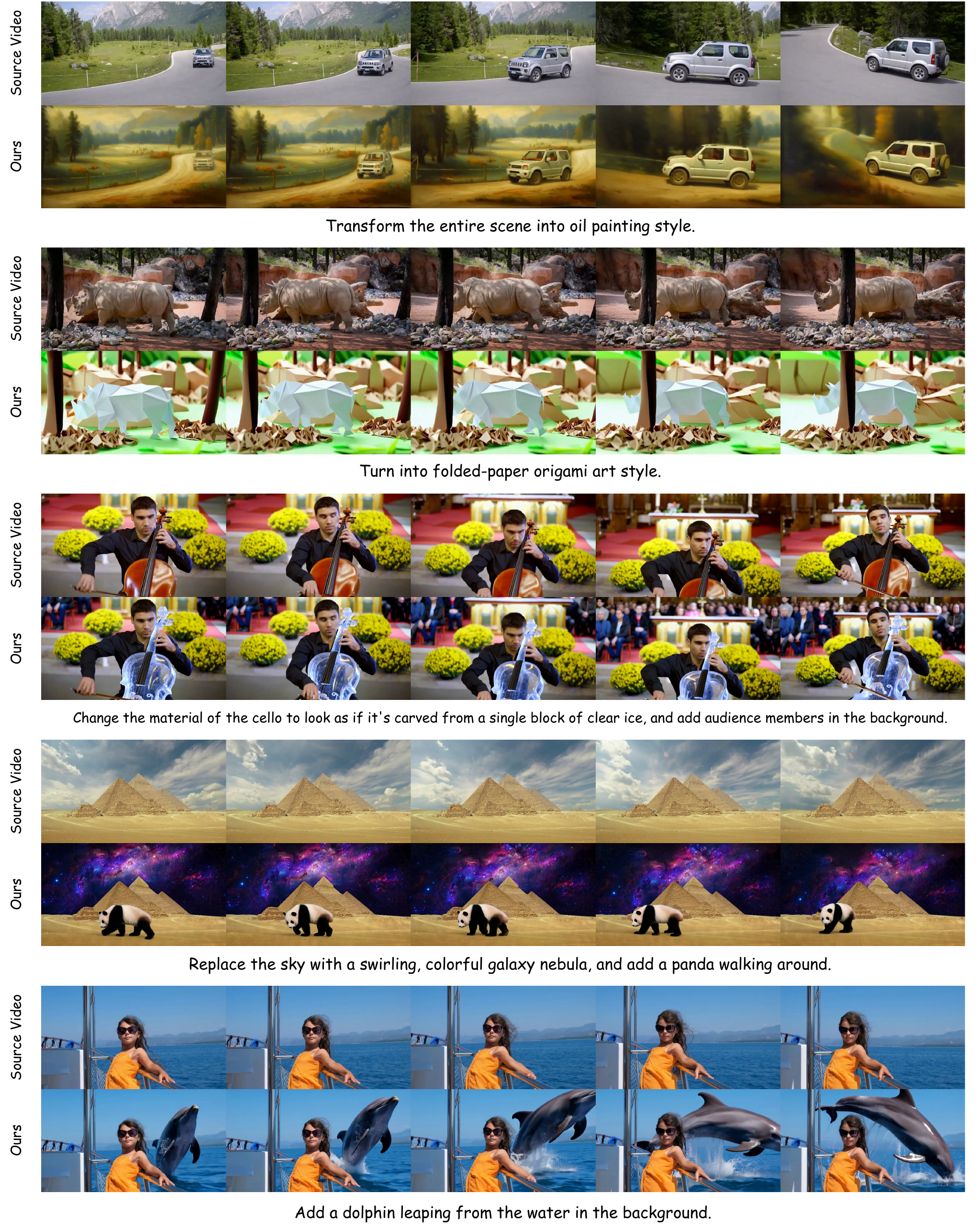} 
\caption{Qualitative results. We present more results of our method on complex instructions that are non-trivial and challenging to be synthesized by the data construction pipeline.
The editing instruction is shown at the bottom.
}
\label{fig:sup:pick_data_1}
\end{figure*}

%% file: figures/fig_sup_qualitative_2.tex
\begin{figure*}
\centering
\includegraphics[width=0.95\textwidth]{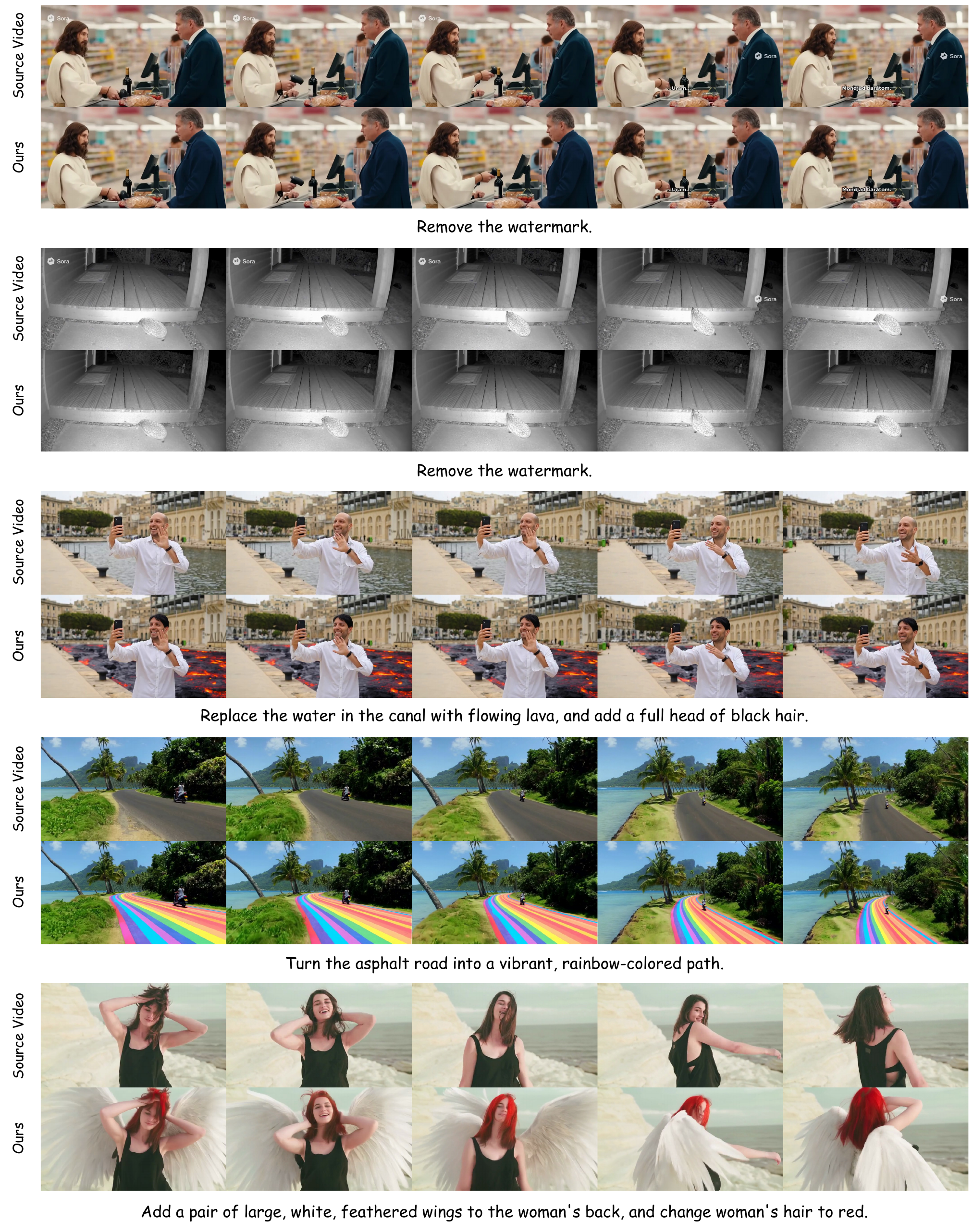} 
\caption{Qualitative results. We present more results of our method on complex instructions that are non-trivial and challenging to be synthesized by the data construction pipeline.
The editing instruction is shown at the bottom.
}
\label{fig:sup:pick_data_2}
\end{figure*}

%% file: figures/fig_sup_vlmtemplate.tex
\begin{figure*}
\centering
\includegraphics[width=1\textwidth]{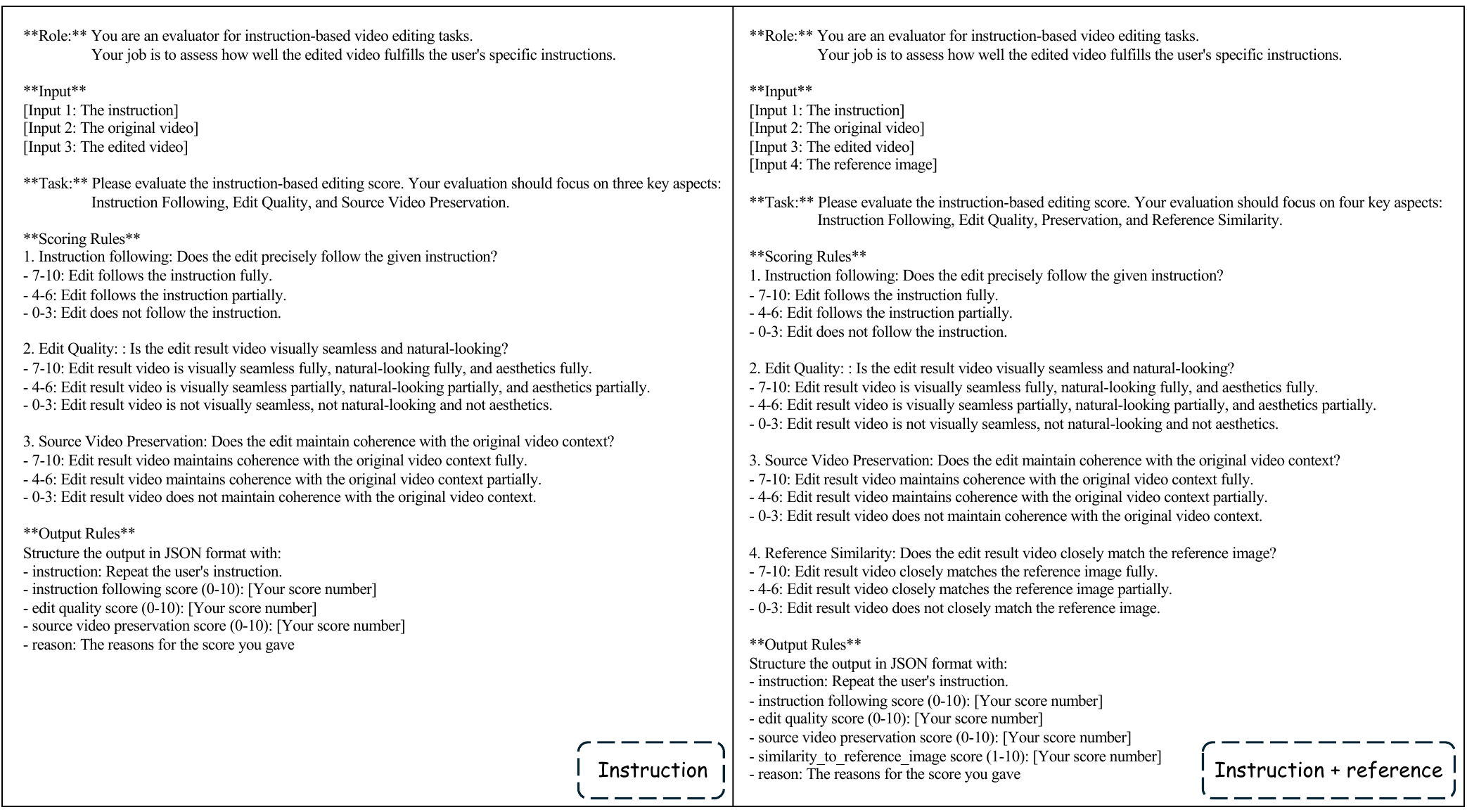} 
\caption{VLM templates for the instruction-based video editing and reference-instruction-based video editing.
}
\label{fig:sup:vlm_template}
\end{figure*}

%% file: appendix/limitation.tex
\section{Limitations}
\label{appendix:sec:limitations}

Figure~\ref{fig:sup:failure_cases} presents three failure cases of \modelname.
Despite the exceptional generalization capabilities of \modelname, it encounters challenges in specific cases.
Rapid motion might occasionally lead to blurry outputs, such as the woman's hand. 
Furthermore, \modelname sometimes struggles to balance editing intensity: it tends to exhibit over-editing in global transformations (such as weather or style changes) while showing under-editing in removal tasks, where residual artifacts—such as cast shadows—often remain.